\documentclass[]{fairmeta}
\usepackage{lmodern}
\usepackage{hyunin}

\DeclareMathOperator{\gca}{\texttt{GCA}} 
 
\DeclareMathOperator{\ca}{\texttt{CA}}
\DeclareMathOperator{\ffn}{\texttt{FFN}}
\DeclareMathOperator{\enc}{\texttt{Enc}}

\usepackage[rgb]{xcolor} 
\definecolor{myblue}{rgb}{0.0,0.498,1.0} 
\definecolor{paperred}{HTML}{d62728} 
\definecolor{paperblue}{HTML}{1f77b4}
\definecolor{lightred}{rgb}{1.0, 0.8, 0.8}

\newboolean{commentver}
\setboolean{commentver}{false}
\ifthenelse{\boolean{commentver}}{
\newcommand{\hyun}[1]{\textcolor{lightred}{\small [#1 --HL]}}
\newcommand{\cj}[1]{\textcolor{orange}{\small [#1 --CJ]}}
}{
\newcommand{\hyun}[1]{{}}
\newcommand{\cj}[1]{{}}  
}

\newcommand{\daggersym}{\text{\dag}}
\usepackage[utf8]{inputenc}

\title{Cross-attention Secretly Performs Orthogonal Alignment in Recommendation Models}

\author[1,2,*]{Hyunin Lee}
\author[1,\daggersym]{Yong Zhang}
\author[1]{Hoang Vu Nguyen}
\author[1]{Xiaoyi Liu}
\author[1]{Namyong Park}
\author[1]{Christopher Jung}
\author[1]{Rong Jin}
\author[1]{Yang Wang}
\author[1]{Zhigang Wang}
\author[2,\daggersym]{Somayeh Sojoudi}
\author[1,\daggersym]{Xue Feng}

\affiliation[1]{Meta}
\affiliation[2]{UC Berkeley}
\contribution[*]{Work done at Meta}
\contribution[\daggersym]{Equally advised}

\abstract{
 
Cross-domain sequential recommendation (CDSR) aims to align heterogeneous user behavior sequences collected from different domains. 
While cross-attention is widely used to enhance alignment and improve recommendation performance, its underlying mechanism is not fully understood.
Most researchers interpret cross-attention as residual alignment, where the output is generated by removing redundant and preserving non-redundant information from the query input by referencing another domain data which is input key and value.
Beyond the prevailing view, we introduce \textbf{Orthogonal Alignment}, a phenomenon in which cross-attention discovers novel information that is not present in the query input, and further argue that those two contrasting alignment mechanisms can co-exist in recommendation models
We find that when the query input and output of cross-attention are orthogonal, model performance improves over 300 experiments.
Notably, Orthogonal Alignment emerges naturally, without any explicit orthogonality constraints. \textbf{Our key insight is that Orthogonal Alignment \emph{emerges naturally} 
because it improves scaling law}. We show that baselines additionally incorporating cross-attention module outperform parameter-matched baselines, achieving a superior accuracy–per–model parameter.
We hope these findings offer new directions for parameter-efficient scaling in multi-modal research.
}

\date{\today}
\correspondence{\email{hyunin@berkeley.edu}, \email{yongzhang@meta.com}}

\begin{document}

\maketitle

\begin{figure}[ht]
    \centering
    \begin{subfigure}[t]{0.32\textwidth}
        \centering
        \includegraphics[width=\textwidth]{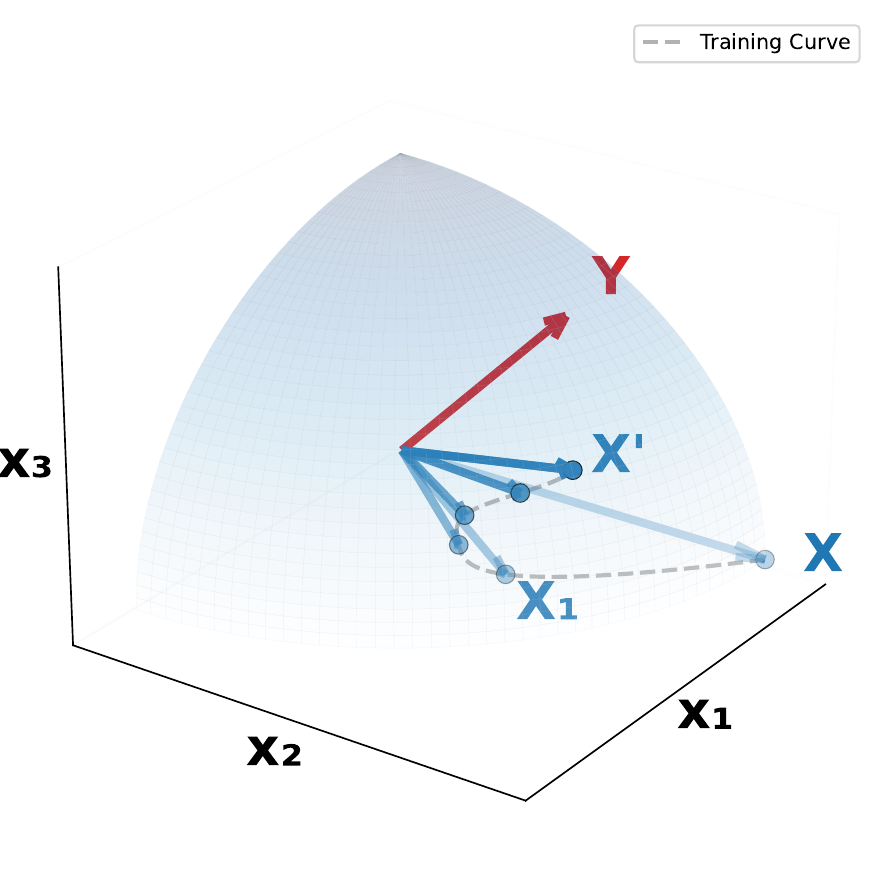}
        \caption{Residual alignment}
        \label{subfig:Parallelism-centric alignment}
    \end{subfigure}
    \begin{subfigure}[t]{0.32\textwidth}
        \centering
        \includegraphics[width=\textwidth]{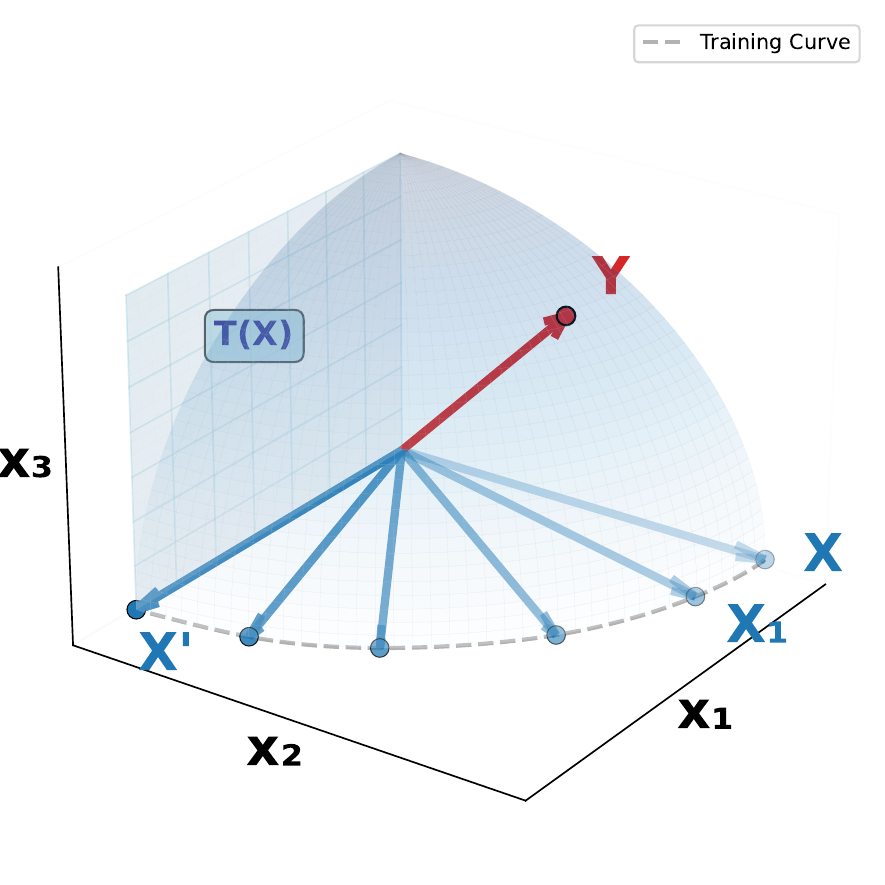}
        \caption{Orthogonal alignment}
        \label{subfig:Orthogonal alignment}
    \end{subfigure}
    \begin{subfigure}[t]{0.2\textwidth}
        \centering
        \includegraphics[width=\textwidth]{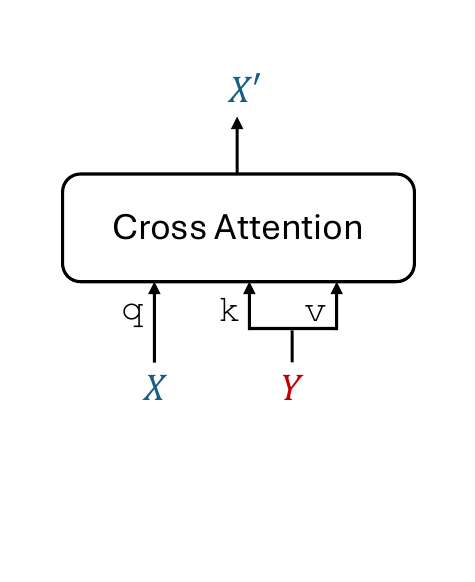}
        \caption{Cross-attention}
        \label{fig:Cross Attention}
    \end{subfigure}
    \caption{ 
    Conceptual illustration of Orthogonal Alignment.
    Given a source representation vector \textcolor{paperred}{$Y$} from domain $B$, suppose the algorithm progressively updates target representation vector \textcolor{paperblue}{$X$} from domain $A$ throughout training iterations $\{X_1, X_2,\cdots, X'\}$.
    (a) Residual alignment: The prevailing view of the cross-attention is that it refines \textcolor{paperblue}{$X$} by reducing irrelevant and preserving relevant information by referring \textcolor{paperred}{$Y$} to update \textcolor{paperblue}{$X'$}. (b) \textbf{Orthogonal Alignment}: We observe a complement-discovery phenomenon where \textcolor{paperblue}{$X'$} becomes increasingly orthogonal to \textcolor{paperblue}{$X$} as model performance improves (Subsection~\ref{subsec:Observation 2}). 
    We show that this orthogonality emerges because cross-attention enables parameter-efficient scaling by extracting complementary information from an orthogonal manifold \textcolor{paperblue}{$T(X)$}, thus enhancing performance without a proportional increase in parameters (Subsection~\ref{subsec:Observation 3}).
    (c) \textcolor{paperblue}{$X'$} is the output of cross-attention, with \textcolor{paperblue}{$X$} as the query and \textcolor{paperred}{$Y$} as the key and value.
    }
    \label{fig:OA}
\end{figure}

\section{Introduction}
\label{sec:intro}
In recent years, the rapid growth in artificial intelligence (AI) has led to an explosion not only in data volume but also in data diversity. Users leave interaction traces across multiple platforms (e.g., Facebook, Instagram, Amazon), within different scenarios on a single platform (e.g., buying products, leaving comments, clicking ads), and even across various categories or boards within a single scenario (e.g., books, movies, groceries). Furthermore, the advent of the transformer architecture \citep{vaswani2017attention}, has significantly advanced recommendation systems, enabling the extraction of user intent from behavioral sequences. 

As a result, cross-domain sequential recommendation (CDSR) systems have emerged, aiming to combine heterogeneous behavioral sequences from diverse sources to improve overall recommendation performance with the hope that the signals collected from various sequential data sources can complement each other. However, naive approaches to combining signals often suffer from performance degradation due to noisy, redundant, or conflicting inter-domain information  \citep{park2023cracking,zhang2023collaborative,li2023one,gong2025multiple} and this has led to one of the main challenges in CDSR: designing an fusion architecture that can effectively handle these heterogeneous sequences \citep{wu2025image,ye2025harnessing,wang2025cross,malitesta2025formalizing} 
\cj{One of the main challenges in cross-domain sequential recommendation system is designing an fusion architecture to handle heterogenous behavioral sequences}\hyun{revised}. 
The most widely adopted solution is the cross-attention mechanism, which aligns and projects representations from different domains into a unified latent space
\citep{liu2021cross,lin2024mixed,ju2025revisiting}. 

Despite its popularity, the internal mechanisms of cross-attention across domains remain poorly understood and are largely explored through empirical studies. Current research views cross-attention as enabling one domain (\textcolor{paperblue}{$X$} in Figure~\ref{fig:Cross Attention}) to query another (\textcolor{paperred}{$Y$} in Figure~\ref{fig:Cross Attention}) and integrate only the most relevant information (\textcolor{paperblue}{$X^\prime$} as a weighted sum of \textcolor{paperred}{$Y$} in Figure~\ref{fig:Cross Attention}). Empirical studies show that cross-attention aligns two sequences by suppressing noise and redundancy, ensuring that only non-redundant information is passed forward. For example, in text-to-image diffusion, cross-attention maps yield faithful token-to-region correspondences that act as denoising and relevance filters rather than blind fusion \citep{tang2023daam, yamaguchi2024mmdit, yang2024crossattn_disentangle}. \cj{What do we mean by visualization?}\hyun{erased} Further analyses indicate that cross-attention serves as an inductive bias, promoting disentanglement of complementary factors and encouraging aligned, non-redundant representations \citep{yang2024crossattn_disentangle}. In vision-language models (VLMs), attention alignment with human gaze further validates that effective cross-attention focuses on causally relevant regions \citep{yan2024voilaa}, while generic transformer interpretability methods \citep{chefer2021beyondattn} provide tools to trace this information flow. Therefore, understanding the cross-attention mechanism as a \emph{residual alignment} is an prevalent interpretation within the research community. 

\hyun{would be great to mention whether multi-modal recsys also interpret cross attention as residual alignment}

This work challenges this conventional view and uncovers a new, counter-intuitive mechanism of cross-attention. We argue that two contrasting alignment mechanisms are able to co-exist within recommendation models. We show that gated cross-attention ($\gca$) improves recommendation performance\cj{Do we want to restrict ourselves to recommendation system or just state this result more broadly?}\hyun{No, but this is an important question that readers in AI field may naturally ask. I addressed in [section \ref{sec:QnAs}-Q4].} 
by producing output that is not merely filtered version of the input query, but is naturally constructed to be \emph{complementary} by exploring previously unseen input query\cj{"are actively constructed" sound a bit awkward because it's a passive voice but it's saying "actively".. I think we can just say ``by producing outputs that are not merely filtered versions of the input but that actively complement it with signals absent from the original input"}\hyun{revised - I use naturally instead of actively. Also, viewing the orthogonal alignment as passive phenomenon is also what I claim and intended.}.
We define this phenomenon as \textbf{Orthogonal Alignment}\cj{Should we give the previous alignment also a name like this in this intro and not just in the figure?}\hyun{revised- named as residual alignment}. Specifically, orthogonal alignment refers to a representational alignment mechanism where the input query (\textcolor{paperblue}{$X$}) and the output (\textcolor{paperblue}{$X^\prime$}) of the cross-attention are orthogonal, rather than simply reinforcing the existing pre-aligned features of \textcolor{paperblue}{$X$} when updating to \textcolor{paperblue}{$X^\prime$} (See Figure~\ref{fig:OA} a visual illustration of this phenomenon, contrasted with the prior residual-alignment).

Crucially, we classify orthogonal alignment as a phenomenon because we empirically show that it \emph{emerges naturally, without requiring ANY explicit orthogonality regularization} in either the loss formulation or model architecture. We observe that $\gca$ intrinsically induces orthogonalization between the input query and the output (Section~\ref{subsec:Observation 2}), leading to significant improvements in ranking performance (Section~\ref{subsec:Observation 1}). Through comprehensive empirical validation -- including over 300 experimental configurations spanning diverse random initializations, architectural variations of the $\gca$ module, and evaluations across three recent CDSR algorithms and four multi-domain dataset combinations -- we establish a robust negative correlation between the cosine similarity metric and the recommendation performance (Section~\ref{subsec:Observation 2}). 

\hyun{may mention that we also use $X$ and its orthogonal part $X'$ as linear combination (residual connection) and that leads to model performance.}

\textbf{Most importantly, our main contribution is explaining why orthogonal alignment naturally occurs. We argue that this phenomenon improves a scaling law.} By ensuring that updates occupy subspace orthogonal to the input query, the model gains new representational capacity without needing more parameters. To demonstrate this, we compare two approaches: adding a $\gca$  module to the baseline model versus simply increasing the baseline model’s parameters. We show that the baseline with the $\gca$ module consistently outperforms the parameter-augmented baseline (see Section~\ref{subsec:Observation 3}). Therefore, we conclude that the ability of cross-attention's inner mechanism to extract orthogonal information is closely linked to the model’s capacity for parameter-efficient scaling. 

To this end, we hope these findings encourage a paradigm shift in both the recommendation and multi-modal learning communities—moving beyond conventional residual alignment metrics toward developing fine-grained orthogonal alignment measures, thereby can advance fundamental approaches for parameter-efficient scaling in multi-modal data fusion.

\section{Related works}

\textbf{Multi-modal Alignment.} Multi-modal alignment was first popularized by models such as CLIP \citep{Radford2021CLIP} and ALIGN \citep{Jia2021ALIGN}, which demonstrated that contrastive objectives can effectively align image and text embeddings in a shared space, enabling strong transferability. These methods train an image encoder and a text encoder jointly by optimizing a contrastive loss: maximizing cosine similarity for matching image–text pairs and minimizing it for non-matching pairs. This approach established the concept of \emph{alignment as parallelism}—encouraging heterogeneous representations to converge in the embedding space.

However, alignment as parallelism is not directly incorporated in baseline models, which typically pre-train image and text encoders separately before integrating them into larger architectures. Additionally, another prominent technique for modality alignment; cross-attention is widely used, but it remains less fully understood. Empirical studies show that cross-attention aligns two sequences by suppressing noise and redundancy, ensuring that only non-redundant information is propagated. For instance, in text-to-image diffusion models, cross-attention maps produce accurate token-to-region correspondences, functioning as denoising and relevance filters rather than indiscriminate fusion mechanisms \citep{tang2023daam, yamaguchi2024mmdit, yang2024crossattn_disentangle}.

Further research indicates that cross-attention acts as an inductive bias, promoting the disentanglement of complementary factors and encouraging the formation of aligned, non-redundant representations \citep{yang2024crossattn_disentangle}. In vision-language models (VLMs), attention alignment with human gaze supports the view that effective cross-attention targets causally relevant regions \citep{yan2024voilaa}, while transformer interpretability methods \citep{chefer2021beyondattn} offer tools to trace this information flow. Thus, understanding cross-attention as a residual-alignment mechanism has become a prevalent interpretation in the research community.

However, recent research has raised concerns about the limitations of \emph{residual-alignment} in multi-modal models. For instance, \cite{Dufumier2024What} argues that strict alignment can restrict learning to only the redundant components shared across modalities, potentially overlooking unique or synergistic modality-specific information. To address this, CoMM was introduced, which aligns representations in a fused multimodal space using a mutual-information objective. This approach allows complementary and unique features to emerge, rather than enforcing redundancy.
Similarly, \cite{Fahim2024Not} explores the phenomenon where image and text embeddings occupy disjoint regions in the joint latent space—a situation known as the modality gap. The study finds that this gap may be an inherent consequence of the contrastive objective itself, referring to it as the contrastive gap. To mitigate this, the authors propose adding uniformity and alignment terms to the loss function.

These findings indicate that enforcing strict residual-alignment may sometimes suppress valuable non-redundant information. All those previous findings motivates a natural research question; can allowing the emergence of complementary (orthogonal) representations enhance the ability of multi-modal models to capture more diverse information across modalities?


\textbf{Cross-domain sequential recommendation.} In recommendation systems, cross-domain sequential recommendation (CDSR) extends above ideas to behavioral sequences of users across heterogeneous item domains.

Conventional sequential recommendation architectures model users' temporal preference dynamics through unimodal behavioral trajectories (e-commerce interactions, video consumption sequences, etc.). Contemporary user engagement patterns, however, manifest across heterogeneous domains with semantically complementary characteristics (e.g., culinary content consumption preceding kitchenware acquisition behaviors). Cross-domain sequential recommendation (CDSR) consequently addresses the fundamental challenge of predicting subsequent items in target domains through unified modeling of chronologically-structured interaction sequences emanating from disparate source domains \citep{zhu2021cross,chen2024survey}.

Early CDSR methodologies concentrated on coarse-grained interest fusion mechanisms. The Mixed Interest Network (MiNet) paradigm models both domain-invariant and domain-specific user preferences through attentive integration of heterogeneous click histories from advertising and news domains, yielding substantial improvements in click-through prediction accuracy \citep{ouyang2020minet}. Subsequently, RecGURU employs adversarial learning principles: a gradient-reversal discriminator constrains the user encoder to learn domain-agnostic representations while preserving recommendation fidelity, demonstrating robust performance gains across short-video and e-commerce benchmarks \citep{li2022recguru}.

The advent of Transformer-based sequential architectures \citep{vaswani2017attention} catalyzed investigations into fine-grained sequential pattern exploitation. Dual Attentive Sequential Learning (DASL) introduces paired self-attention mechanisms to simultaneously capture domain-specific and cross-domain dependencies, surpassing prior fusion baselines across multiple evaluation datasets \citep{chen2021dual}. The Mixed Attention Network (MAN) further decomposes local (domain-specific) and global (cross-domain) signals through hierarchical attention architectures and item-similarity fusion mechanisms, achieving state-of-the-art performance without requiring user overlap constraints \citep{lin2024mixed}. Complementarily, Hiformer addresses heterogeneous feature interaction challenges in web-scale recommender systems through novel Transformer-based architectures with heterogeneous self-attention layers, optimizing both feature interaction learning and inference efficiency \citep{gui2023hiformer}. Beyond deterministic encoding paradigms, meta-learning and stochastic modeling approaches have garnered considerable attention. CDSR with Neural Processes (CDSRNP) conceptualizes CDSR as a stochastic process and leverages neural process frameworks to learn distributional representations over user sequences, facilitating rapid domain adaptation with limited behavioral observations \citep{li2024cdrnp}. Complementarily, Triple-sequence Learning with CDSR (Tri-CDR) proposes triple sequence modeling—concurrently representing source, target, and hybrid sequences—to disentangle transferable knowledge from domain-specific preferences through contrastive alignment mechanisms \citep{Ma2023TripleSL}.

Recent developments have embraced large language model (LLM) integration and advanced architectural innovations. LLM4CDSR leverages the powerful representational and reasoning capabilities of large language models to address the overlap dilemma and transition complexity inherent in CDSR, introducing LLM-based unified representation modules with trainable adapters and contrastive regularization to bridge semantic item relationships \citep{liu2025bridge}. Meanwhile, ABXI introduces novel cross-perturbation mechanisms for enhanced cross-domain knowledge transfer, employing sophisticated data augmentation strategies to improve generalization across heterogeneous domains \citep{bian2025abxi}.

Collectively, these methodological contributions trace an evolutionary trajectory from rudimentary mixed-interest fusion (MiNet), through adversarial and dual-attention transfer mechanisms (RecGURU, DASL), to sophisticated attention decomposition strategies (MAN, Hiformer), LLM-enhanced semantic reasoning (LLM4CDSR), advanced perturbation techniques (ABXI), and meta-level sequence modeling paradigms (CDSRNP, Tri-CDR).
\section{Experiment}
\label{sec:Experiment}
\begin{figure}[h!]
    \centering
    \includegraphics[width=0.9\linewidth]{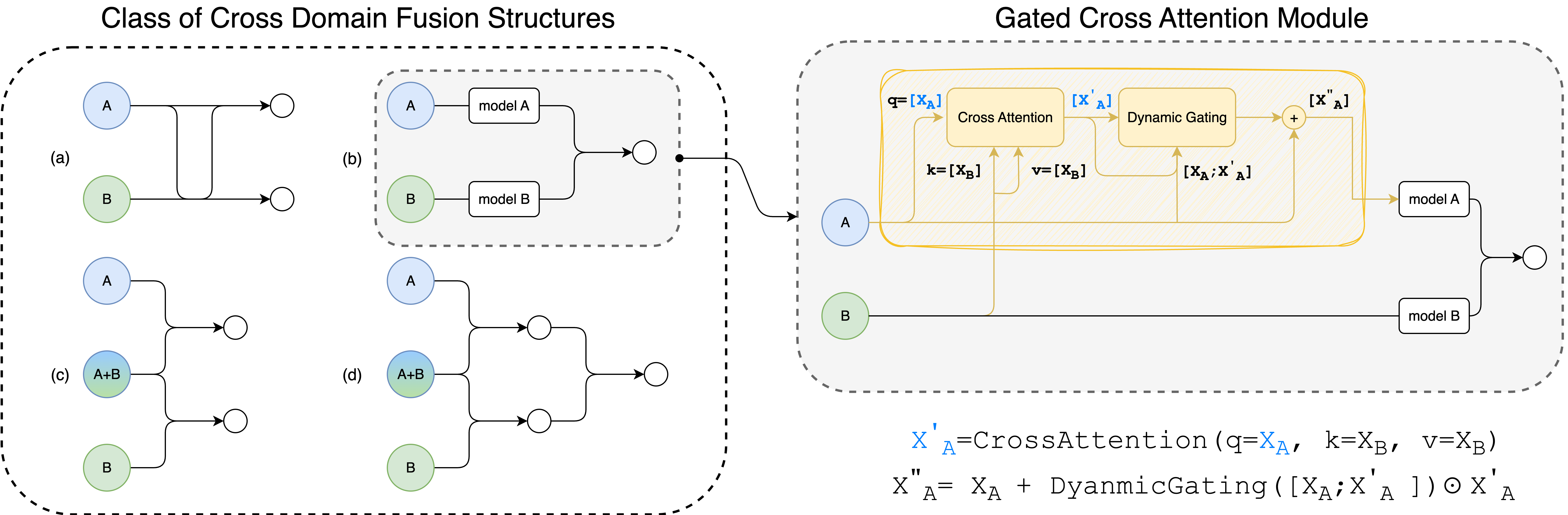}
    \caption{In cross-domain sequential recommendation, various fusion structure ((a) $\sim$ (d)) across heterogeneous sequence data forms the central backbone of the model. In this work, we propose a gated cross-attention mechanism applied at the early interaction stage between two domain sequences. Empirical results show that this module consistently improves recommendation performance (see Section \ref{subsec:Observation 1}). Our main analysis reveals  that a primary role of gated cross-attention is to induce orthogonal representations of the query inputs (see Section \ref{subsec:Observation 2}). Specifically, we observe that a reduction in cosine similarity between \textcolor{myblue}{$X_A$} and its cross-attended counterpart \textcolor{myblue}{$X_A'$} correlates strongly with enhanced recommendation accuracy.}
    \label{fig:algorithm}
\end{figure}

We first explain the gated cross-attention ($\gca$) module (Figure \ref{fig:algorithm}) then explain how we utilized $\gca$ on top of the baseline model.

\subsection{Gated Cross Attention Module}
\label{subsec:}

Let $X \in \mathbb{R}^{B \times l_X \times d}$ denote a item feature sequence with batch size $B$, sequence length $l_X$, and item feature dimension $d$. For two domains $A$ and $B$, we define $X_A \in \mathbb{R}^{B \times l_A \times d}$ and $X_B \in \mathbb{R}^{B \times l_B \times d}$ as the item sequences from domains $A$ and $B$, respectively, with which the same user has interacted. We denote $\ffn([X_A, X_B])$ as a sequence-wise and batch-wise feedforward network that operates on the concatenated representation $[X_A; X_B] \in \mathbb{R}^{B \times \max(l_A, l_B) \times 2d}$ and outputs a representation in $\mathbb{R}^d$. We denote $\ca(\text{query}=X_A,~ \text{key}=X_B,~ \text{value}=X_B)$ as a multi-head cross-attention mechanism with $X_A$ as the query and $X_B$ as both key and value. For simplicity, we denote it as $\ca(X_A, X_B)$ or equivalently $\ca(\text{q}=X_A,~\text{k,v}=X_B)$.  

Then,the $\gca$ module is then formulated as
\[
\gca(X_A, X_B) = \texttt{Layernorm}(X_A + \ffn([X_A; X_B]) \odot X_A'),
\]  
where $X_A' = \ca(X_A, X_B)$ and $\odot$ denotes the Hadamard (element-wise) product and $\texttt{Layernorm}$ is a layernorm operator \citep{ba2016layer}. In this formulation, $\ffn([X_A; X_B])$ serves as a gating mechanism that produces dimension-wise gating values. These gating values control the degree to which information from the cross-attended representation $X_A'$ is incorporated into the original representation $X_A$, thereby enabling the model to selectively propagate informative signals. In practice, $\ffn([X;Y])$ is implemented as a two-layer feedforward neural network with either a $\texttt{sigmoid}$ or $\texttt{tanh}$ activation function at the output layer.

Our design is inspired by the Flamingo model~\citep{alayrac2022flamingo}, an early visin language model,  but differs in a key aspect: instead of employing a fixed gating structure, our gating module learns vector-valued gating outputs conditioned on the concatenated input sequences. To the best of our knowledge, this is the first work to introduce a gated cross-attention module in the context of recommendation research. 

\subsection{\texorpdfstring{$\gca$}{GCA}s on Top of Baselines}


In the context of cross-domain sequential recommendation, we selected recent baseline models that follow a two-step design principle: (1) treating multiple domain sequences individually, rather than concatenating them into a single sequence at the early stage (i.e., each domain sequence is regarded as a distinct feature source); and (2) applying self-attention to the concatenated feature embeddings to capture higher-order dependencies.  

The three algorithms considered here represent recent advances that have demonstrated superior performance compared to conventional cross-domain sequential recommendation methods such as \texttt{Pi-Net}, \texttt{DASL}, \texttt{C}$^2$\texttt{DSR}, \texttt{DREAM}, and \texttt{TriCDR}. To the best of our knowledge, prior work has not investigated whether augmenting such baselines with a gated cross-attention (GCA) module could further improve performance. Moreover, no prior study has examined whether enhancing the orthogonality between the output of cross-attention and the input query contributes to improved recommendation accuracy.  

\begin{figure}[h]
    \centering
    \includegraphics[width=0.95\linewidth]{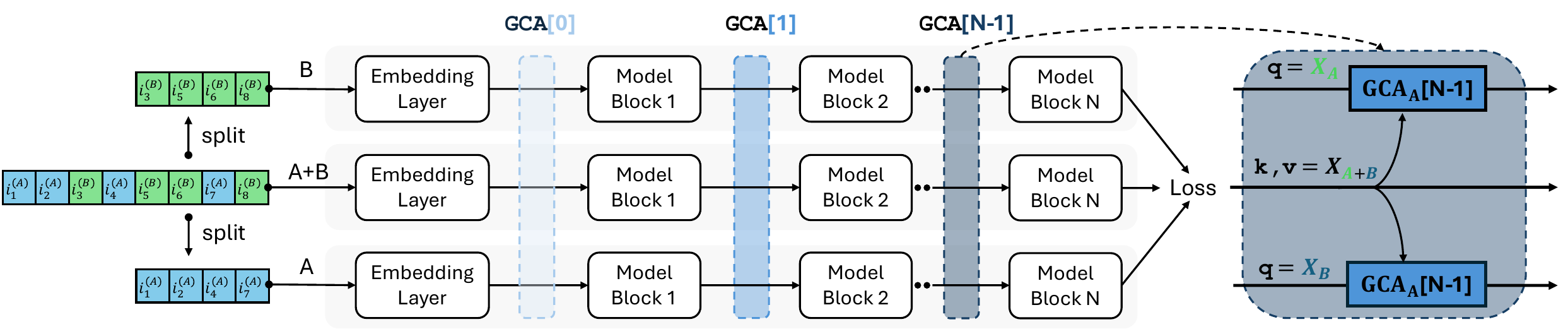}
    \caption{For each baseline model, we insert $\gca$ modules at multiple vertical positions, denoted as $\gca[i]$, where $i=0$ corresponds to the module closest to the raw data and $i=N$ to the module farthest from the raw data. By design, $\gca[0]$ is always placed immediately after the embedding layer, while $\gca[1], \gca[2], \ldots$ are positioned within intermediate layers of the backbone. Each $\gca[i]$ comprises two parallel gated cross-attention modules, which respectively refine the representations of domains $A$ and $B$.}
    \label{fig:GCA_architecutre}
\end{figure}

A common characteristic of these baselines is that the architecture begins by splitting a mixed sequence $X_{A+B} \in \mathbb{R}^{B \times (l_A + l_B) \times d}$ into two domain-specific sequences, $X_A \in \mathbb{R}^{B \times l_A \times d}$ and $X_B \in \mathbb{R}^{B \times l_B \times d}$, where $\{X_A\} \cup \{X_B\} = \{X_{A+B}\}$ and $\{X_A\} \cap \{X_B\} = \emptyset$. With $X_{A+B}, X_A,$ and $X_B$ as inputs, a shared (ABXI) or independent (CDSRNP, LLM4CDSR) Transformer-style self-attention encoder is employed to learn sequential dependencies in user behavior. We denote this encoder as $\enc$, which consists of multiple stacked multi-head self-attention layers.  

Figure~\ref{fig:GCA_architecutre} illustrates how additional $\gca$ modules are integrated into the baseline architectures at different positions. We denote $\gca_{i}[n]$ as a GCA module with type $i \in \{A, B\}$ and vertical position index $n \in \{0,1,\cdots, N\}$. Specifically, $\gca_{A}[n]$ uses $X_A$ as the query and either $X_B$ or $X_{A+B}$ as the key and value, while $\gca_{B}[n]$ uses $X_B$ as the query and either $X_A$ or $X_{A+B}$ as the key and value. For a fixed $i$, the index $n$ indicates the vertical placement of the GCA module within the network: smaller values of $n$ correspond to earlier stages (e.g., closer to the embedding layer), while larger values correspond to deeper layers (e.g., closer to the final representation). In summary, the index $i$ specifies the domain alignment (i.e., how many GCAs are placed in parallel at the same depth), whereas $n$ specifies the vertical position of the GCA module within the overall architecture, from input embeddings to the final representation used for loss computation. In the experiment, we set $N=2$.

We next describe the three baseline models (CDSRBP, ABXI, and LLM4CDSR) together with their $\gca$-augmented baseline
.



\textbf{CDSRNP.}~\citep{li2024cdrnp}\footnote{\url{https://github.com/cjx96/CDSRNP}}  
The CDSRNP model formulates cross-domain sequential recommendation as a conditional neural process (CNP) to transfer user preference information across domains. For overlapped users, behavior sequences are partitioned into a support set $\mathcal{S} = \{(X^{(s)}_u, Y^{(s)}_u)\}$ and a query set $\mathcal{Q} = \{(X^{(q)}_u, Y^{(q)}_u)\}$. The encoder maps $\mathcal{S}$ into latent representations $r_s = \texttt{Enc}(X^{(s)}_u, Y^{(s)}_u)$, which are aggregated into a global context $c = \frac{1}{|\mathcal{S}|}\sum_s r_s$. For a query sequence $X^{(q)}$, the decoder produces predictions $\hat{Y}^{(q)} = \texttt{Dec}(X^{(q)}, c, z)$, where $z \sim q_\phi(z|\mathcal{S},\mathcal{Q})$ is sampled from the approximate posterior. The prior distribution $p_\theta(z|\mathcal{S})$ is aligned with $q_\phi$ through KL divergence regularization. To capture fine-grained user-specific interests, an adaptive refinement layer modifies query embeddings as $X^{(q)}_{\text{adapt}} = X^{(q)} \odot \sigma(Wc)$, where $W$ is a learnable projection. This design allows CDSRNP to infer cross-domain correlations even for non-overlapped users by leveraging latent function inference from the neural process framework.  

\begin{itemize}
    \item $\text{raw semantic data} \xrightarrow{\text{Embedding}_i} X_{i} \xrightarrow{\gca_i[0]} \texttt{Enc}_i(X_i) \xrightarrow{\gca_i[1]} \cdots \to \text{final representation}, \quad i \in \{A, B, A+B\}$
\end{itemize}

In this architecture, the thread corresponding to $X_{A+B}$ plays an auxiliary role: CDSRNP concatenates $\texttt{Enc}(X_{A+B})$ with $\texttt{Enc}(X_A)$ and $\texttt{Enc}(X_B)$, thereby constructing augmented representations $[\texttt{Enc}(X_A); \texttt{Enc}(X_{A+B})]$ and $[\texttt{Enc}(X_B); \texttt{Enc}(X_{A+B})]$, which serve as enriched features for downstream analysis.




\textbf{ABXI.}~\citep{bian2025abxi}\footnote{\url{https://github.com/DiMarzioBian/ABXI}}  
The ABXI model is designed for task-guided cross-domain sequential recommendation. Its central objective is to extract invariant user interests that transfer effectively across domains, while simultaneously allowing each domain to maintain domain-specific specialization. The architecture begins by splitting a mixed sequence ($X_{A+B}$) into two domain-specific sequences ($X_A$ and $X_B$). With $X_{A+B}, X_A,$ and $X_B$ as inputs, a shared encoder with multiple self-attention layers learns sequential dependencies in user behavior. On top of this backbone, ABXI incorporates Low-Rank Adaptation (LoRA)-based domain adapters—lightweight low-rank adaptation layers—that fine-tune the shared representation for each specific domain.  

\begin{itemize}
    \item raw semantic data $\xrightarrow{\text{Embedding}_i} X_{i} \xrightarrow{\gca_i[0]} \texttt{dropout}(X_{i}) \xrightarrow{\gca_i[1]} \enc(\texttt{dropout}(X_{i})) = X_{i}^{(1)}, \quad i \in \{A,B, A+B\}$
    \item $X_{i}^{(1)} + \texttt{DLORA}_{i}(X_{i}^{(1)}) = X_{i}^{(2)} \xrightarrow{\gca_i[2]} X_{i}^{(2)} + \texttt{ILORA}_{i}(X_{A+B}^{(2)}) \to \cdots \to \text{final representation}, \qquad i \in \{A,B\}$
\end{itemize}

Note that ABXI uses shared $\enc$ different from LLM4CDSR or CDSNRP. We explore three possible placements of gated cross-attention (GCA) within ABXI, inserted vertically across the architecture. Each single $\gca_i$ ($i \in \{0,1,2\}$) consists of two parallel GCAs: $\gca(\text{q}=X_A, \text{k,v}=X_{A+B})$ and $\gca(\text{q}=X_B, \text{k,v}=X_{A+B})$, which update the domain-specific representations $X_A$ and $X_B$ by referencing the mixed-domain sequence $X_{A+B}$. Since our study focuses on early interactions between sequences, we design more fine-grained options at Step~1, applying GCA either immediately after the raw input sequence ($\gca_0$) or after the dropout layer ($\gca_1$), both prior to the multi-head self-attention module. Each placement offers trade-offs: applying GCA before dropout regularizes the input representations but may lead to information loss, whereas applying it after dropout preserves input quality but provides less regularization, thereby reducing robustness to noise and increasing the risk of overfitting.

\textbf{LLM4CDSR.}~\citep{liu2025bridge}\footnote{\url{https://github.com/Applied-Machine-Learning-Lab/LLM4CDSR-pytorch}} The LLM4CDSR model employs a tri-thread sequential recommendation architecture that jointly captures domain-specific preferences and cross-domain behavioral patterns. The three threads process $X_{A+B}$, $X_A$, and $X_B$, respectively, with each consisting of an embedding layer followed by a multi-head self-attention module. For the cross-domain thread ($X_{A+B}$), the embedding layer is initialized with pre-trained LLM-based text embeddings, which are frozen during training to preserve the semantic knowledge encoded in the original LLM representations. By contrast, the two local domain threads adopt trainable embedding layers, initialized via PCA-transformed LLM embeddings derived from the frozen $X_{A+B}$ embeddings, to enhance stability during training. We formalize the baseline model, along with the insertion points of the proposed gated cross-attention (GCA) modules, as follows:  
\begin{itemize}
    \item $\text{raw semantic data} \xrightarrow{\text{Embedding}_i} X_{i} \xrightarrow{\gca_i[0]} \enc_i(X_i) \xrightarrow{\gca_i[1]}\cdots \to \text{final representation}$
\end{itemize}
In this formulation, $\gca_0$ is applied immediately after the embedding layer, whereas $\gca_1$ is inserted after multiple layers of $\texttt{MHA}$. This architecture provides a strong and competitive baseline by combining the semantic richness of LLM-based embeddings with explicit modeling of both local and cross-domain sequential dynamics, thereby establishing a robust foundation for evaluating the effectiveness of the proposed gated cross-attention mechanism. Note that $\text{Embedding}_A,~\text{Embedding}_B$ are intialized by frozen $\text{Embedding}_{A+B}$.

\textbf{Dataset}. We utilized the public dataset Amazon Reviews \citep{hou2024bridging} where the domains are different product types (such as book, movie, sport, music, etc). Since this work is investigating the performance increase and the role of $\gca$ adhoc model on top of public baseline algorithms, we take variety on different cross domain combinations such as Cloth-Sport, Electronic-Phone, Food-Kitchen, etc. 

\section{Experiment Discussion}
\label{sec:Experiment Discussion}
Extensive experiments reveal four observations, summarized as follows. Our main contributions are Observations 2 and 3 (Subsections \ref{subsec:Observation 2} and \ref{subsec:Observation 3}).

\subsection{Observation 1: \texorpdfstring{$\gca$}{GCA} at the Early Stage Consistently Improves Performance (NDCG, AUC).}
\label{subsec:Observation 1}

We evaluate the effect of gated cross-attention ($\gca$) by comparing three backbones with and without the module: LLM4CDSR vs. LLM4CDSR+$\gca$, ABXI vs. ABXI+$\gca$, and CDSRNP vs. CDSRNP+$\gca$. Table \ref{Table1.Experiment results} reports NDCG and AUC on both domains $A$ and $B$. Here, $\gca_\text{early}$ corresponds to $\gca[0]$, where the module is inserted at the first interaction stage. $\gca_\text{stack}$ denotes deeper variants such as ${\gca[0,1], \gca[0,1,2], \ldots}$, where $\gca[0]$ is combined with additional vertically stacked $\gca[i]$ modules ($i > 1$). Among several stacked configurations (e.g., $\gca[0,1]$, $\gca[0,2]$), we report the best test performance. Again, each $\gca[i]$ consists of two parallel $\gca$ blocks: $\gca_A[i]$ and $\gca_B[i]$.

Specifically, for LLM4CDSR and ABXI, $\gca$ is applied as $\gca(\text{q} = X_A, \text{k,v} = X_{A+B})$ and $\gca(\text{q} = X_B, \text{k,v} = X_{A+B})$ where we use combined sequence $X_{A+B}$ as a query, and for CDSRNP, $\gca$ is applied in a pairwise fashion as $\gca(q = X_A, k,v = X_B)$ and $\gca(q = X_B, k,v = X_A)$. During the training, we retain baseline learning hyperparameters (learning rate, decay, hidden dimension, attention heads, etc) that was proposed from the baseline paper across both baselines and baseline$+\gca$ models. In Table \ref{Table1.Experiment results}, we report NDCG@1, NDCG@10, and AUC, and all reported performance metrics are averaged over five seeds. 

\begin{table}[H]
\centering
\small
\setlength{\tabcolsep}{4pt}
\resizebox{\textwidth}{!}{%
\begin{tabular}{lccccccc}
\toprule
Model & Dataset (A-B) & NDCG@1$_\text{A}$ & NDCG@10$_\text{A}$ & NDCG@1$_\text{B}$ & NDCG@10$_\text{B}$ & AUC$_\text{A}$ & AUC$_\text{B}$ \\
\midrule
\midrule
LLM4CDSR & Cloth-Sport & 0.7157{\tiny$\pm$0.0025} & 0.7821\tiny{$\pm$0.0018} & 0.5870 \tiny {$\pm$ 0.0051} & 0.6493 \tiny{$\pm$ 0.002} & 0.9216 {\tiny $\pm$0.0013} & 0.8621 {\tiny$\pm$ 0.0054} \\

$+\gca_{\text{early}}$ &  &  0.7283{\tiny$\pm$ 0.0027} & 0.8052\tiny{$\pm$0.0014} & 0.5977 \tiny {$\pm$ 0.0054} &  0.6560 \tiny{$\pm$ 0.0046} & 0.9364 {\tiny $\pm$0.0009} & 0.8655 {\tiny$\pm$ 0.0038} \\

$+\gca_{\text{stack}}$ &  &  0.7310{\tiny$\pm$ 0.0012} & 0.8056\tiny{$\pm$0.0014} &  0.6112 \tiny {$\pm$ 0.0032} &   0.6638 \tiny{$\pm$ 0.0038} &  0.9370 {\tiny $\pm$0.0010} & 0.8664 {\tiny$\pm$ 0.0030} \\

\midrule

LLM4CDSR & Elec-Phone & 0.2101{\tiny$\pm$0.0030} & 0.3512\tiny{$\pm$0.0009} & 0.1419 \tiny {$\pm$ 0.0008} & 0.2608 \tiny{$\pm$ 0.0010} & 0.7901 {\tiny $\pm$0.0008} & 0.7197 {\tiny$\pm$ 0.0011} \\

$+\gca_{\text{early}}$ &  &  0.2378{\tiny$\pm$ 0.0011} & 0.3815\tiny{$\pm$0.0018} & 0.1861 \tiny {$\pm$ 0.0035} &  0.2845 \tiny{$\pm$ 0.0027} & 0.7970 {\tiny $\pm$0.0018} & 0.7218 {\tiny$\pm$ 0.0026} \\

$+\gca_{\text{stack}}$ &  &  0.2410{\tiny$\pm$ 0.0012} & 0.3800\tiny{$\pm$0.0011} &  0.1994 \tiny {$\pm$ 0.0054} &   0.3035 \tiny{$\pm$ 0.0049} &  0.7937 {\tiny $\pm$0.0013} & 0.7252 {\tiny$\pm$ 0.0026} \\

\midrule
\midrule

ABXI & Beauty-Elec &  0.0730{\tiny$\pm$0.0070} & 0.1724\tiny{$\pm$0.0071} & 0.0548 \tiny {$\pm$ 0.0038} & 0.1273 \tiny{$\pm$ 0.0028} &  0.7216 {\tiny $\pm$0.0027} & 0.7123 {\tiny$\pm$ 0.0009} \\

$+\gca_{\text{early}}$ &  &  0.0727{\tiny$\pm$ 0.0060} & 0.1793\tiny{$\pm$0.0047} & 0.0544 \tiny {$\pm$ 0.0044} &  0.1244 \tiny{$\pm$ 0.0025} & 0.7410 {\tiny $\pm$0.0025} & 0.7169 {\tiny$\pm$ 0.0024} \\

$+\gca_{\text{stack}}$ &  &  0.0733{\tiny$\pm$ 0.0042} & 0.1846\tiny{$\pm$0.0057} &  0.0566 \tiny {$\pm$ 0.0052} &   0.1271 \tiny{$\pm$ 0.0042} &  0.7354 {\tiny $\pm$0.0048} & 0.6973 {\tiny$\pm$ 0.0051} \\

\midrule 

ABXI & Food-Kitch &  0.0593{\tiny$\pm$0.0074} & 0.1541\tiny{$\pm$0.0130} &  0.0416 \tiny {$\pm$ 0.0058} & 0.1093 \tiny{$\pm$ 0.0113} &  0.7205 {\tiny $\pm$ 0.0015} & 0.7180 {\tiny$\pm$  0.0032} \\

$+\gca_{\text{early}}$ &  &  0.0703{\tiny$\pm$ 0.0094} & 0.1757\tiny{$\pm$0.0092} & 0.0548 \tiny {$\pm$ 0.0053} &  0.1327 \tiny{$\pm$ 0.0072} & 0.7317 {\tiny $\pm$0.0039} & 0.7150 {\tiny$\pm$ 0.0031} \\

$+\gca_{\text{stack}}$ &  &  0.0882{\tiny$\pm$ 0.0052} & 0.1853\tiny{$\pm$0.0013} &  0.0527 \tiny {$\pm$ 0.0020} &   0.1282 \tiny{$\pm$ 0.0028} &  0.7148 {\tiny $\pm$0.0026} & 0.6924 {\tiny$\pm$ 0.0009} \\

\midrule 
\midrule

CDSRNP & Elec-Phone (1M) &  0.0499 {\tiny$\pm$0.0087} & 0.1170\tiny{$\pm$0.0079} &  0.0920 \tiny {$\pm$ 0.0050} & 0.1935 \tiny{$\pm$ 0.0021} &  - & - \\

$+\gca_{\text{early}}$ &  & 0.0547{\tiny$\pm$ 0.0010} & 0.1209 \tiny{$\pm$0.0092} & 0.0980 \tiny {$\pm$ 0.0010} &  0.1989 \tiny{$\pm$ 0.0031} & - & - \\

$+\gca_{\text{stack}}$ &  &  0.0531{\tiny$\pm$ 0.0078} & 0.1229\tiny{$\pm$0.0125} &  0.0946\tiny {$\pm$ 0.0022} &   0.1942 \tiny{$\pm$ 0.0012} &  - & - \\

\bottomrule
\end{tabular}
}
\caption{NCDG and AUC comparison with the three baselines and adhoc model with GCA. Elec stands for Electronic, and Kitch stands for Kitchen. $\gca_\text{early}$ denotes $\gca[0]$ and $\gca_{\text{stack}}$ denotes $\gca[0,i_1,i_2,..,i_N]$ where $i_n >1, n \in [N]$}
\label{Table1.Experiment results}
\end{table}

Across all three backbones, introducing an early $\gca$ module ($\gca_{\text{early}}$) improves recommendation accuracy, though the magnitude of gains varies by dataset and architecture. 
\begin{itemize}
    \item \textbf{LLM4CDSR}. On Cloth–Sport dataset, $\gca_{\text{early}}$ increases NDCG@1$_A$ from $0.716$ to $0.728$ ($+1.2$ points) and NDCG@10$_A$ from $0.782$ to $0.805$ ($+2.3$ points), alongside an AUC gain of $+1.5$ points, and also observe gain in Domain $B$. Vertically stacking ($\gca_{\text{stack}}$) yields further gains, pushing NDCG@1$_A$ to $0.731$ and NDCG@1$_B$ from $0.597$ to $0.611$. On Elec–Phone, early $\gca$ improves NDCG@1 by $+2.8$ points on Domain A and $+4.4$ points on Domain B, with modest AUC gains. Therefore, stacking provides additional lift on Domain B (NDCG@1: $0.199$ vs. $0.142$ baseline), indicating that deeper cross-domain coupling can be beneficial in this setting.
    \item \textbf{ABXI}. On Beauty–Elec, early $\gca$ raises AUC consistently ($0.722$ to $0.741$ on Domain A), while stacking slightly improves NDCG@10 but reduces AUC$_B$ ($0.712$ to $0.697$). On Food–Kitchen, early $\gca$ improves both NDCG@1 ($0.059$ to $0.072$) and NDCG@10 ($0.154$ to $0.176$), whereas stacking amplifies NDCG@1 further ($0.081$) but again reduces AUC$_B$. These mixed results suggest that stacking introduces a trade-off: higher top-rank precision (NDCG@1) at the cost of overall discrimination (AUC).
    \item \textbf{CDSRNP.} On Elec–Phone (1M), early $\gca$ provides small but consistent improvements (e.g., NDCG@1 from $0.0499$ to $0.0547$, NDCG@10 from $0.117$ to $0.121$). Stacking yields comparable list-level metrics but notably boosts NDCG@1 on Domain B ($0.0092$ to $0.095$).
\end{itemize}

Overall, $\gca_{\text{early}}$ consistently improves performance across all backbones and datasets, underscoring its effect at the initial alignment stage.

\begin{figure}[h!]
    \centering
    \begin{subfigure}[t]{0.4\textwidth}
        \centering
        \includegraphics[width=\textwidth]{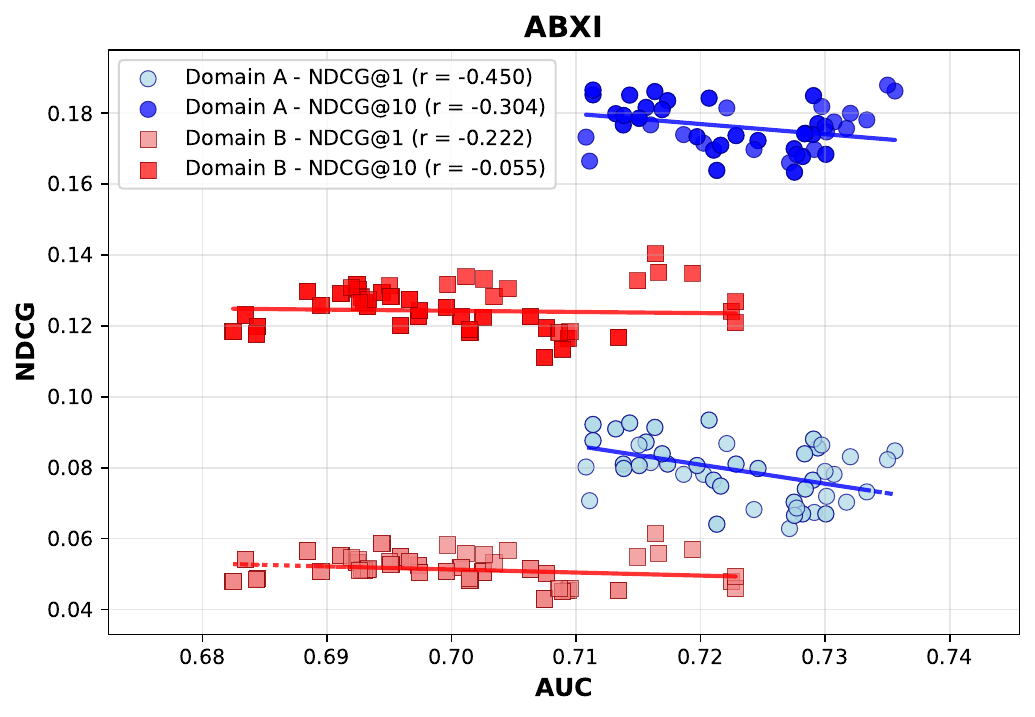}
        \caption{ABXI}
        \label{fig:subfigA}
    \end{subfigure}
    \begin{subfigure}[t]{0.4\textwidth}
        \centering
        \includegraphics[width=\textwidth]{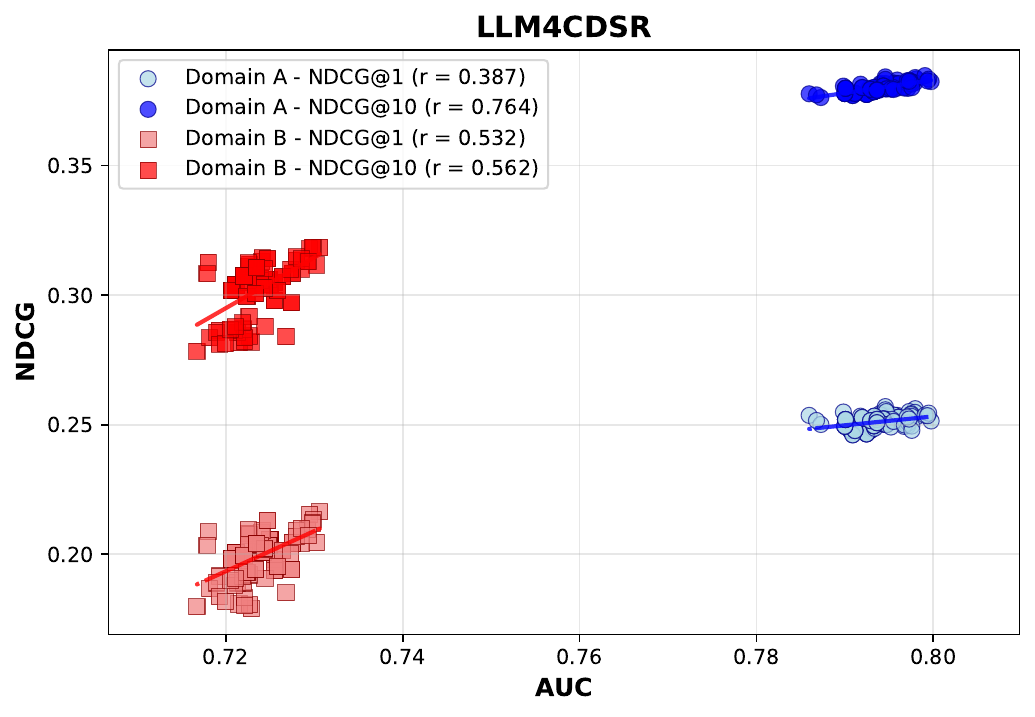}
        \caption{LLM4CDSR}
        \label{fig:subfigB}
    \end{subfigure}
    \caption{NDCG@$\{1,10\}$–AUC correlations differ by backbone: ABXI exhibits a consistent negative correlation across domains, while LLM4CDSR shows a consistent positive correlation.}
    \label{fig:NDCG_vs_AUC}
\end{figure}

\textbf{Trade-off between ranking precision (AUC) and discrimnination (NDCG).} Adding stacked modules can further amplify certain metrics (notably NDCG@1), but the effects are non-monotonic and dataset-dependent, often sharpening head-rank accuracy while degrading AUC (such as both AUC A and B drop for ABXI for both Beaty-Elec and Food-Kitchen dataset). This may raise an question that whether the NDCG and AUC are counter-balancing metric; and to examine this trade-off more closely, we analyze the relationship between NDCG@{1,10} and AUC (See Figure \ref{fig:NDCG_vs_AUC}). Figure \ref{fig:NDCG_vs_AUC} reveals contrasting patterns across models: ABXI exhibits a stable weak negative correlation ($r \in \{ -0.45,-0.30,-0.22,-0.05\}$) between NDCG and AUC in both domains $A$ and $B$, whereas LLM4CDSR shows a consistent strong positive correlation ($r \in \{ 0.38,0.76,0.53,0.56\}$). 

These findings suggest that the interaction between ranking precision (AUC) and discrimination (NDCG) is architecture-specific. These results further highlight the importance of careful placement: $\gca$ is effective as an early-stage alignment mechanism, while deeper stacking should be applied judiciously to avoid undesirable trade-offs.

\begin{figure}[h!]
    \centering
    \begin{subfigure}[t]{0.45\textwidth}
        \centering
        \includegraphics[width=\textwidth]{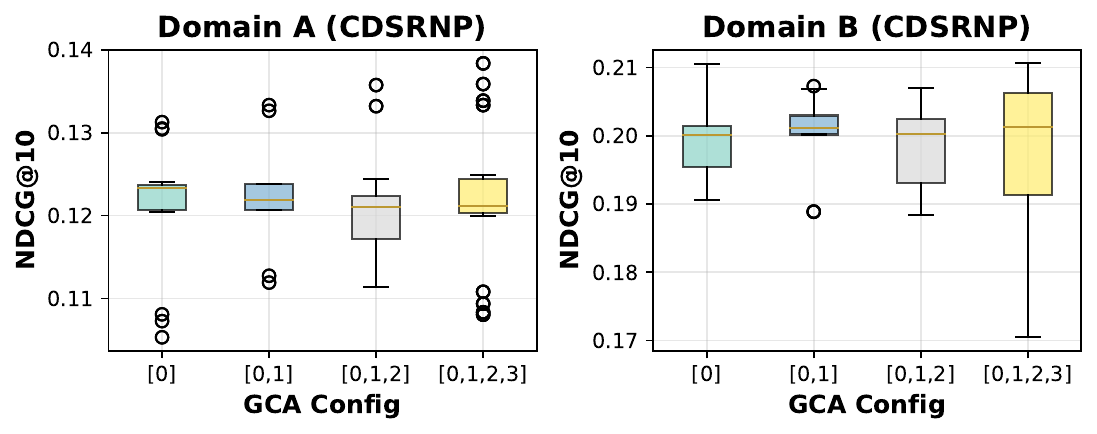}
        \caption{CDSRNP}
        \label{fig:NDCG_vs_GCAconfig_CDSRNP}
    \end{subfigure}
    \begin{subfigure}[t]{0.45\textwidth}
        \centering
        \includegraphics[width=\textwidth]{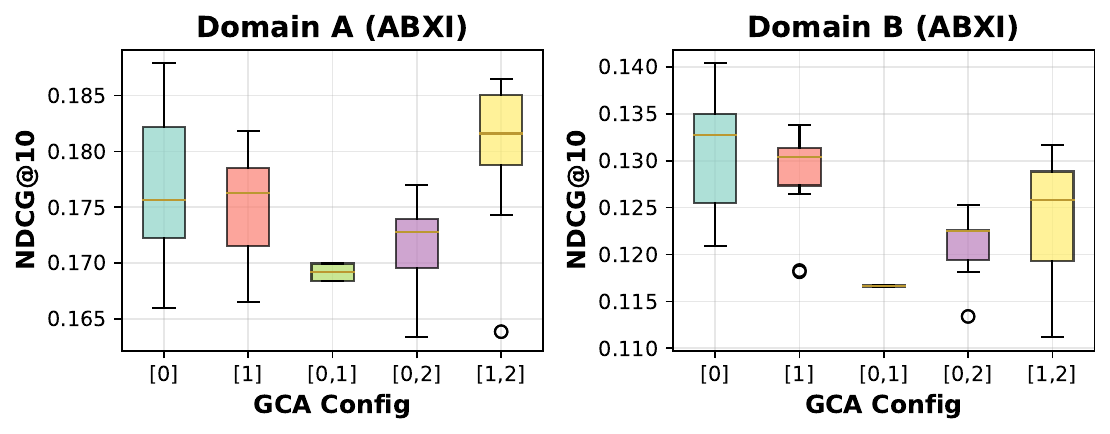}
        \caption{ABXI}
        \label{fig:NDCG_vs_GCAconfig_ABXI}
    \end{subfigure}
    \begin{subfigure}[t]{0.45\textwidth}
        \centering
        \includegraphics[width=\textwidth]{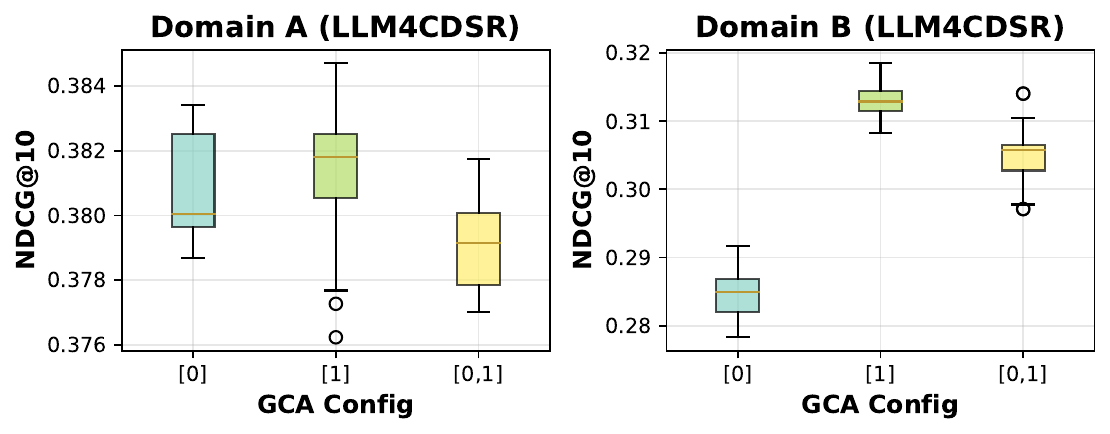}
        \caption{LLM4CDSR}
        \label{fig:NDCG_vs_GCAconfig_LLM4CDSR}
    \end{subfigure}
    \caption{Effect of vertically stacking $\gca$ modules. Increasing the number of insertions (i.e. insert more $\gca[i]\text{s}, i>1$) does not yield monotonic gains: LLM4CDSR peaks at [1], CDSRNP saturates beyond [0,1], and ABXI achieves its best median NDCG at [1,2] but suffers negative transfer with other placements.}
    \label{fig:NDCG_vs_GCAconfig}
\end{figure}

\textbf{Vertically stacking $\gca$s in the baseline is not scalable.} In Table~\ref{Table1.Experiment results}, several cases illustrate that $\gca_\text{stack}$ can underperform relative to $\gca_\text{early}$. We further investigate this and observe that vertically stacking multiple $\gca$ modules \emph{does not necessarily lead to better performance}. Figure \ref{fig:NDCG_vs_GCAconfig} presents a box plot where increasing the number of insertions—from a single placement [0] to deeper stacks like [0,1] or [0,1,2]—does not yield monotonic gains across three backbones (LLM4CDSR, CDSRNP, and ABXI). The orange line in each box indicates the median. 
\begin{itemize}
    \item \textbf{CDSRNP}. Monotonically stacking $\gca$s from [0,1] to deeper stacks like [0,1,2,3] yields no further gains and instead increases variance across runs, pointing to diminishing returns and reduced training stability (see Figure \ref{fig:NDCG_vs_GCAconfig_CDSRNP}).
    \item \textbf{ABXI}. The strongest median NDCG is observed with the selective placement [1,2], while other multi-spot settings underperform—and in some cases fall below the baseline [0]—indicating the negative transfer when $\gca$s are stacked (see Figure \ref{fig:NDCG_vs_GCAconfig_ABXI}).
    \item \textbf{LLM4CDSR}. Inserting $\gca$ only at [1] consistently surpasses the stacked configuration [0,1], suggesting that additional layers may introduce redundancy or disrupt early representations (see Figure \ref{fig:NDCG_vs_GCAconfig_LLM4CDSR}).
\end{itemize}  

This finding aligns with previous research showing that stacking more transformer layers does not guarantee better performance. For instance, in computer vision, CaiT reports ``\emph{no evidence that depth can bring any benefit when training on ImageNet only},'' absent special training strategies \citep{touvron2021going}. Similarly, DeepViT observes that ViT performance saturates quickly when scaled to be deeper due to attention collapse \citep{zhou2021deepvit}. In natural language processing, analogous evidence comes from early-exiting approaches such as DeeBERT, which demonstrates that many transformer layers are effectively redundant and can be skipped during inference while preserving accuracy, saving up to $\sim$40\% inference time with minimal degradation \citep{xin2020deebert}.

Taken together,  the effectiveness of $\gca$ depends strongly on placement and backbone architecture. Stacking more modules or covering more positions does not ensure improvement and can sometimes degrade performance.

Motivated by these findings, the next subsection takes a closer look at the role of $\gca$ and investigate how $\gca$ improves the ranking performance.

\subsection{Observation 2 (Main): \texorpdfstring{$\gca$}{GCA} extracts orthogonal information in a consistent manner regardless of dataset and baseline algorithms.}
\label{subsec:Observation 2}

\begin{figure}[h]
    \centering
    \begin{subfigure}[t]{0.32\textwidth}
        \centering
        \includegraphics[width=\textwidth]{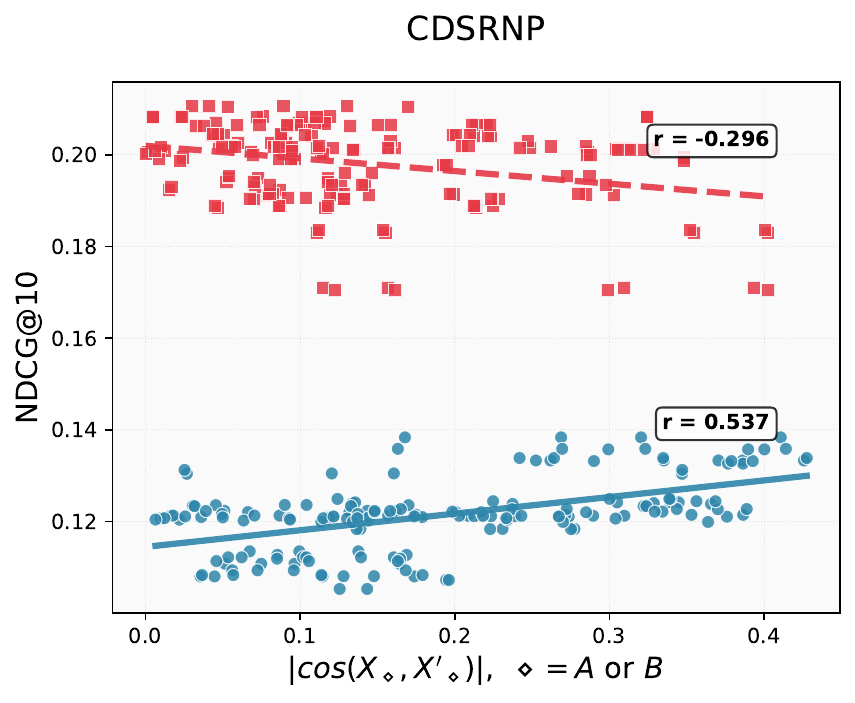}
        \caption{CDSRNP}
        \label{fig:NDCG10_cosXX'_CDSRNP}
    \end{subfigure}
    \begin{subfigure}[t]{0.32\textwidth}
        \centering
        \includegraphics[width=\textwidth]{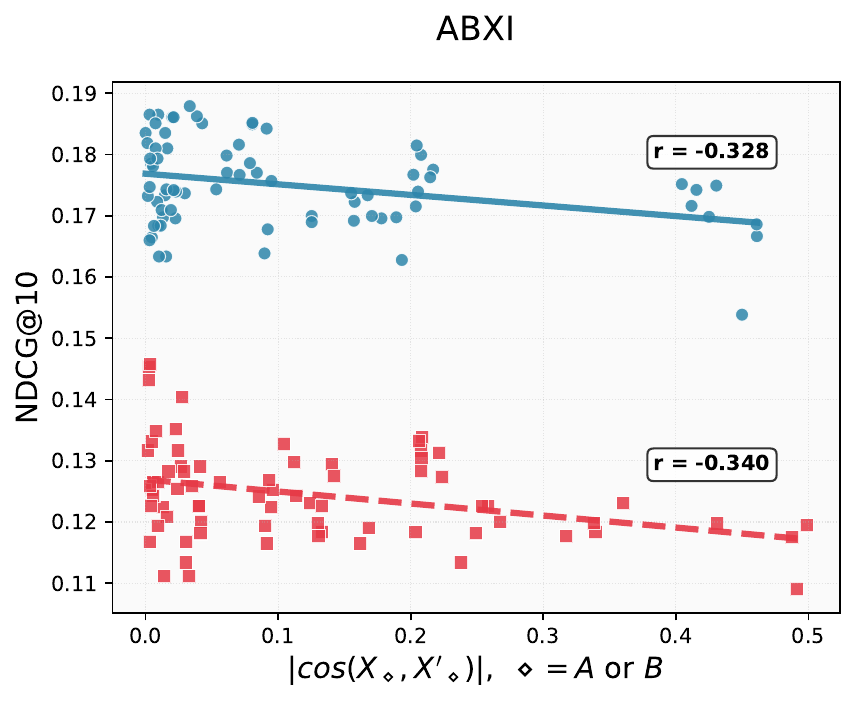}
        \caption{ABXI}
        \label{fig:NDCG10_cosXX'_ABXI}
    \end{subfigure}
    \begin{subfigure}[t]{0.32\textwidth}
        \centering
        \includegraphics[width=\textwidth]{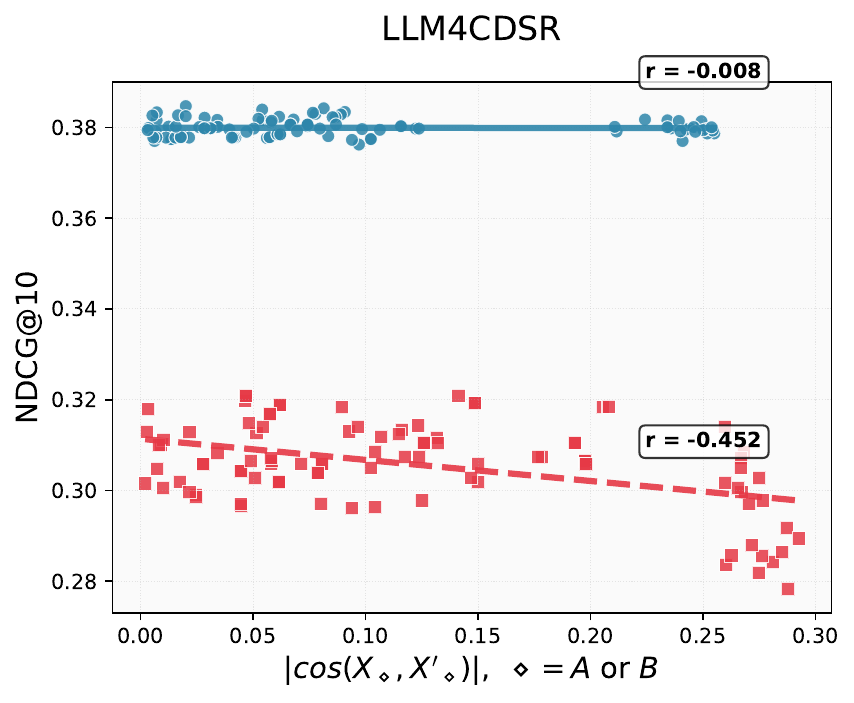}
        \caption{LLM4CDSR}
        \label{fig:NDCG10_cosXX'_LLM4CDSR}
    \end{subfigure}
    \caption{\textbf{Main contribution.} We observed that the gated cross-attention module introduces an unseen, orthogonal feature representation: as the input query $X$ and its cross-attended output $X'$ (conditioned on key and value $Y$) become more orthogonal, the ranking performance improves.}
    \label{fig:NDCG10_cosXX'}
\end{figure}

We identify one key role of $\gca$ as functioning as an intrinsic orthogonal representation regularizer, leading to the Orthogonal Alignment phenomenon. Importantly, the term \emph{orthogonal representation regularizer} should not be interpreted as an explicitly engineered constraint within the $\gca$ module. Rather, it emphasizes that the orthogonal phenomenon \emph{emerges naturally}, without the need for manually imposed regularization. We elaborate on this emergent behavior in Subsection~\ref{subsec:Observation 3}. 

In this subsection, we present the correlation between ranking performance metrics and the degree of orthogonality measured between the cross-attention input (query) and output (updated query).

To evaluate this, we plotted NDCG@10 against $|\cos(X,X')|$ for both domains $A$ and $B$ in Figure \ref{fig:NDCG10_cosXX'}. Here, $|\cos(X,X')|$ represents the batch- and position-averaged cosine similarity between the input query $X_{\diamond}$ and its cross-attended output $X'_{\diamond}$ where $\diamond \in \{ A,B \}$. Formally, for sequences $X_{\diamond}, X'_{\diamond} \in \mathbb{R}^{B \times l{\diamond} \times d}$, with batch index $b \in [B]$ and position $i \in l_{\diamond}$, we computed $|\cos(X,X')| := \frac{1}{Bl_{\diamond}} \sum_{b,i \in [B] \times [l_{\diamond}]} \cos (\vec{X}_{bi}, \vec{X'}_{bi})$ where $\vec{X}_{bi}, \vec{X'}_{bi} \in R^d$ and $\diamond \in \{ A,B\}$. Figure \ref{fig:NDCG10_cosXX'} shows that the correlation between NDCG@10 and $|\cos(X,X')|$ is predominantly negative across domains and backbones, reinforcing the view that $\gca$ enhances recommendation performance by reducing representational overlap between $X$ and $X'$. To be specific, 

\begin{itemize}
    \item \textbf{LLM4CDSR} and \textbf{ABXI}. For LLMCSR, Domain $B$ shows a strong negative correlation ($r=-0.452$), while Domain $A$ shows near-zero correlation ($r=-0.008$). For ABXI, both domains show negative correlations (Domain A: $r=-0.328$, Domain B: $r=-0.340$). Here, stronger orthogonalization consistently aligns with higher NDCG@10, indicating that reducing overlapping information between $X$ and $X'$ is central to performance. 
    \item \textbf{CDSRNP}. The pattern is asymmetric. Domain A shows a positive correlation ($r=0.537$), while Domain B remains negative ($r=-0.296$). This suggests that excessive orthogonalization may suppress useful shared structure, especially in domains with sparse or weak signals, leading to divergent effects.
\end{itemize}

These findings demonstrate that $\gca$ acts not just as an additional attention block, but as a structural constraint that naturally encourages orthogonality between queries and their cross-attended output. This emergent phenomenon is central to its impact on model performance.

\subsection{Observation3 (main): \texorpdfstring{$\gca$}{GCA} is an parameter-efficient way to scale up the model.}
\label{subsec:Observation 3}

\begin{figure}[h!]
    \centering
    \begin{subfigure}[t]{0.25\textwidth}
        \centering
        \includegraphics[width=\textwidth]{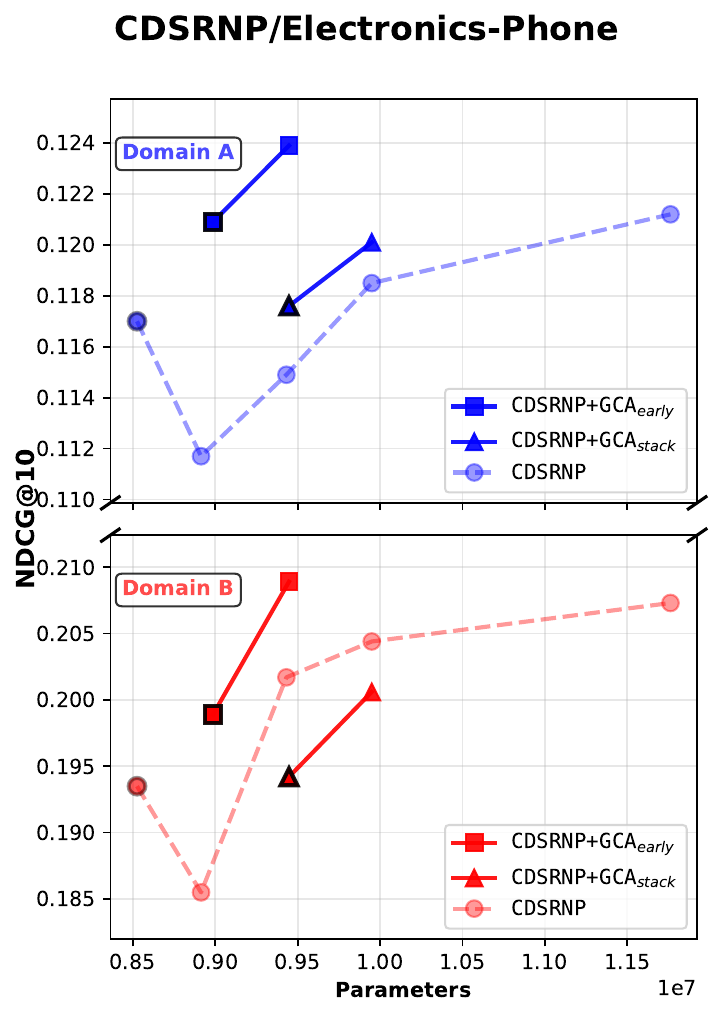}
        \caption{CDSRNP}
        \label{fig:CDSRNP_elec-phone_scaling}
    \end{subfigure}
    \begin{subfigure}[t]{0.25\textwidth}
        \centering
        \includegraphics[width=\textwidth]{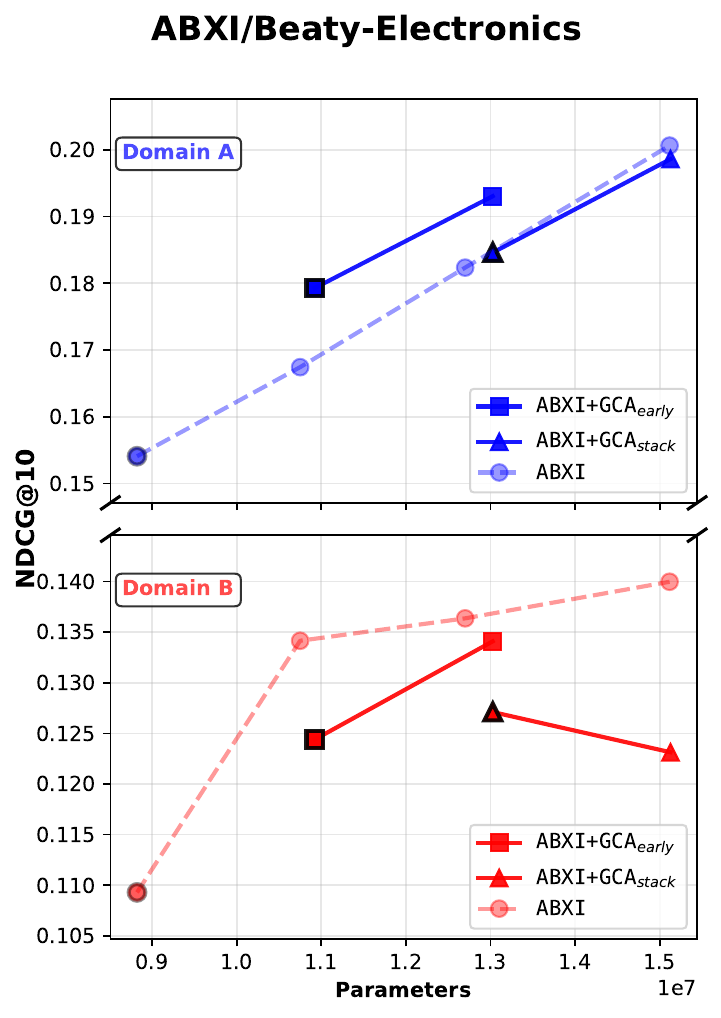}
        \caption{ABXI (Beaty-Elec)}
        \label{fig:ABXI_abe_scaling}
    \end{subfigure}
    \begin{subfigure}[t]{0.25\textwidth}
        \centering
        \includegraphics[width=\textwidth]{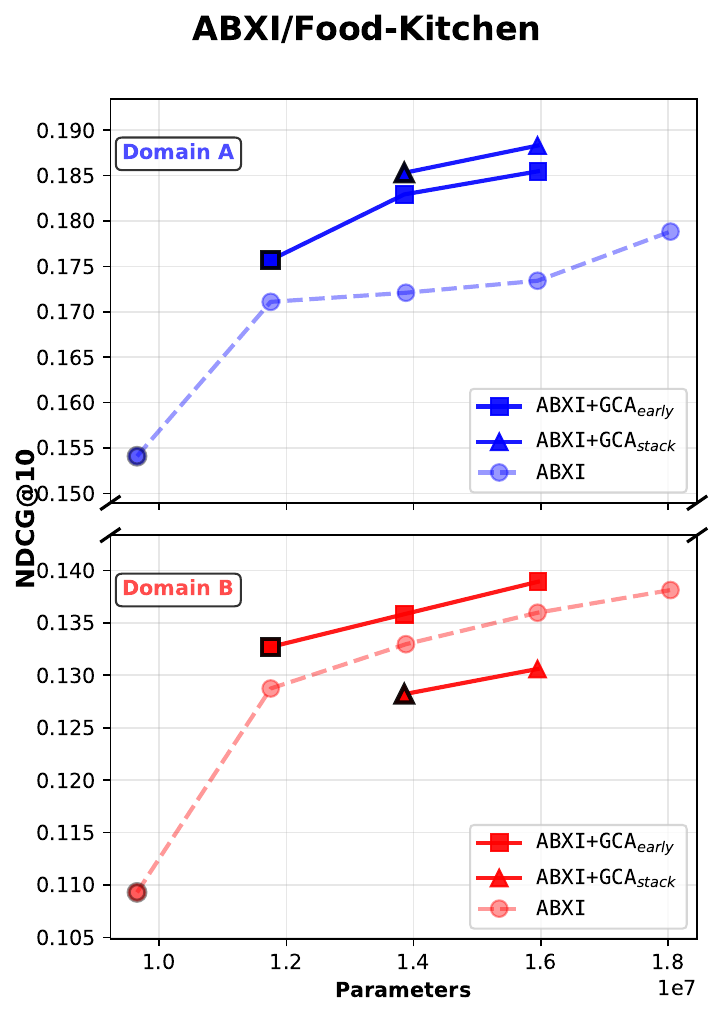}
        \caption{ABXI (Food-Kitch)}
        \label{fig:ABXI_afk_scaling}
    \end{subfigure}
    \begin{subfigure}[t]{0.25\textwidth}
        \centering
        \includegraphics[width=\textwidth]{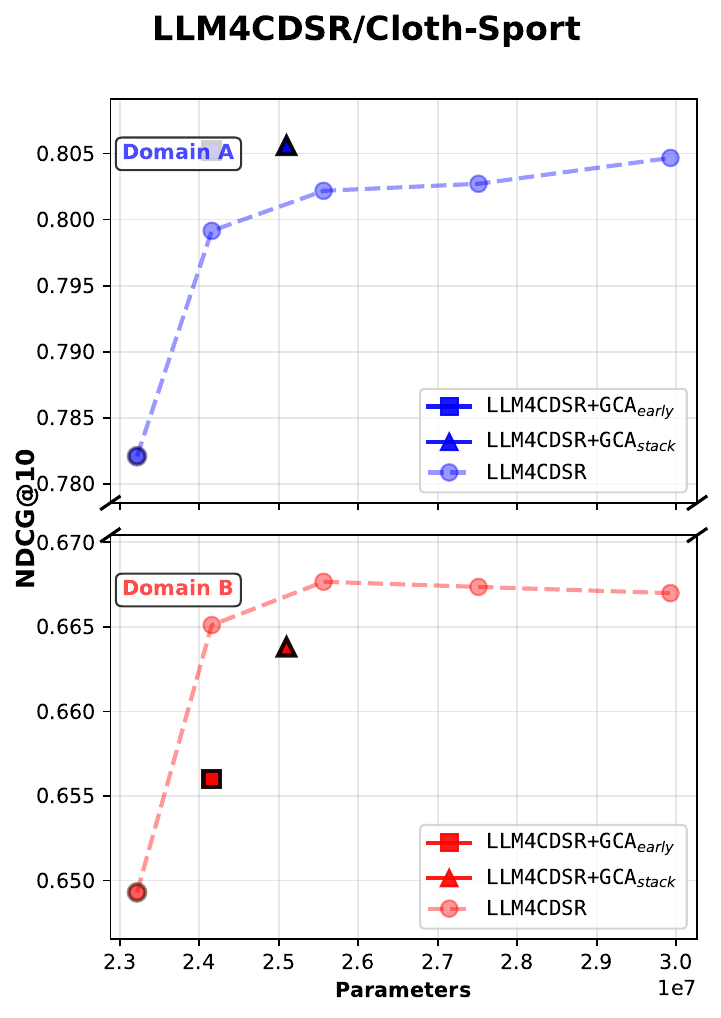}
        \caption{LLM4CDSR (Cloth-Sports)}
        \label{fig:LLM4CDSr_cs_scaling}
    \end{subfigure}
    \begin{subfigure}[t]{0.25\textwidth}
        \centering
        \includegraphics[width=\textwidth]{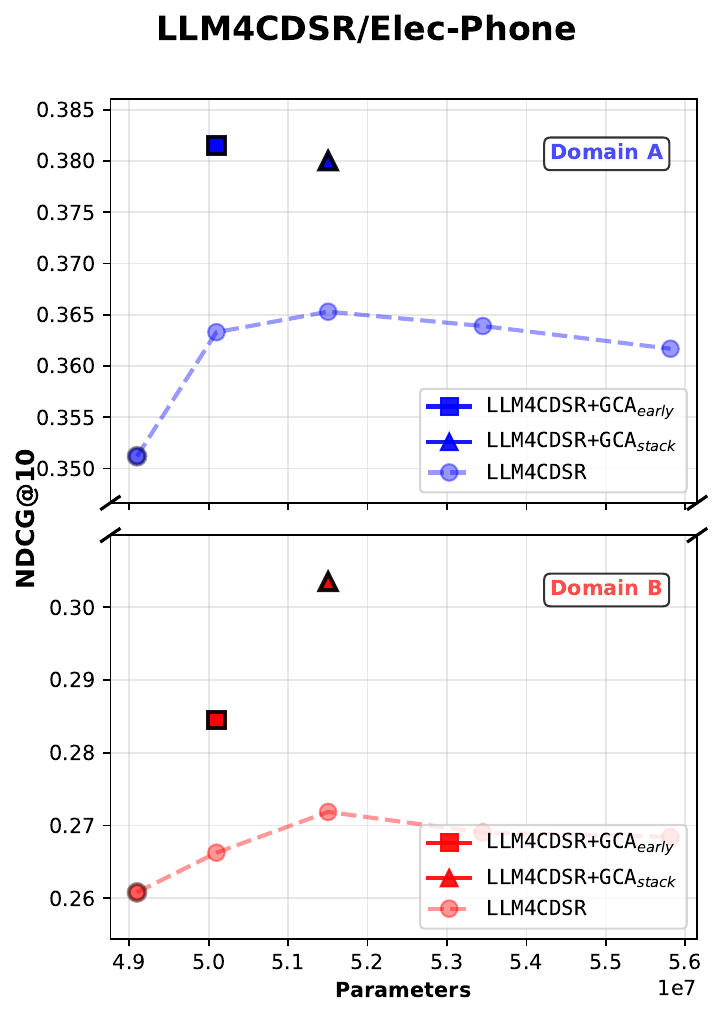}
        \caption{LLM4CDSR (Elec-Phone)}
        \label{fig:LLM4CDSR_ep_scaling}
    \end{subfigure}
    \caption{NDCG comparistion between baseline and adhoc model with $\gca$. There exists three dots that have black edge in each subplots and those three dots are results that reported in Table \ref{Table1.Experiment results}}
    \label{fig:scale up}
\end{figure}

Now, an important question arises: given that the model architecture and loss function do not include \emph{any explicit regularization} terms, how does the observed orthogonality phenomenon \emph{naturally emerge} in Subsection \ref{subsec:Observation 2}? We hypothesize that this phenomenon arises because $\gca$ provides a parameter-efficient means of scaling baseline models; this experimentally substantiates our claim about Orthogonal Alignment. Before presenting the results, it is important to clarify that scaling efficiency may not be the sole explanation for this phenomenon; rather, we view it as one plausible interpretation with other contributing factors requiring further investigation.

To test this hypothesis, we compare the performance of $\gca_{\text{early}}$ and $\gca_{\text{stack}}$ against parameter-matched baselines, thereby controlling for model size. Figure~\ref{fig:scale up} reports the results across the five cases extending the experiments in Table~\ref{Table1.Experiment results}. In each subfigure, three data points with black edges (triangle, rectangle, and circle) correspond to the NDCG@10 and AUC scores from Table~\ref{Table1.Experiment results}, and data points without black edges are extended experiments to check this parameter efficiency hypothesis. For CDSRNP and ABXI, both baselines and their two ad hoc variants ($\gca_{\text{early}}$, $\gca_{\text{stack}}$) are scaled up, whereas for LLM4CDSR only the baseline is scaled. All reported NDCG@10 values of data points are means over five random seeds, where in each run we record the best test NDCG@10. Standard deviations are omitted since they are consistently small and do not affect the conclusions.

First, the results show that for all five cases, $\text{baseline}+\gca_{\text{early}}$ achieves superior single-domain ranking performance (Domain A’s NDCG@10) compared to parameter-matched baselines, with Domain B’s NDCG@10 also generally improved. 

Also, in both LLM4CDSR settings, $\gca_{\text{early}}$ exhibits the strongest parameter efficiency (Figures~\ref{fig:LLM4CDSr_cs_scaling} and \ref{fig:LLM4CDSR_ep_scaling}). We attribute this to the fixed hidden dimensionality of the initial embedding vectors inherited from the pretrained LLM, which imposes an upper bound on naive baseline scaling. Consequently, scaling the baseline alone eventually saturates and even degrades performance as parameter counts increase. In contrast, orthogonal alignment via $\gca$ enables more effective information extraction when the representational capacity of the input space is limited, thereby providing an accuracy-per-parameter advantage.

\subsection{Observation4: \texorpdfstring{$\gca$}{gca} induces orthogonalization independently of how similar \texorpdfstring{$X$}{X} and \texorpdfstring{$Y$}{Y} happen to be.}
\label{subsec:Observation 4}

A natural follow-up from Subsection~\ref{subsec:Observation 2} is whether the statement "reducing $|\cos(X, X')|$ leads to improved performance" depends on the similarity between $X$ and $Y$, measured by $|\cos(X, Y)|$. In other words, does the ability of $X'$ to discover orthogonal information from $X$ depend on how similar the query ($X$) and the key,value ($Y$) are?

To address this, Figure~\ref{fig:cosXY_cosXX'} compares $|\cos(X, X')|$ and $|\cos(X, Y)|$ across three backbones. The boxplots reveal several consistent patterns:

\begin{figure}[h!]
    \centering
    \begin{subfigure}[t]{0.27\textwidth}
        \centering
        \includegraphics[width=\textwidth]{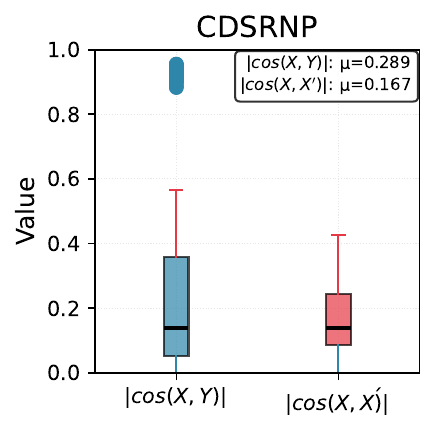}
        \caption{CDSRNP}
        \label{fig:cos_cdsrnp}
    \end{subfigure}
    \begin{subfigure}[t]{0.27\textwidth}
        \centering
        \includegraphics[width=\textwidth]{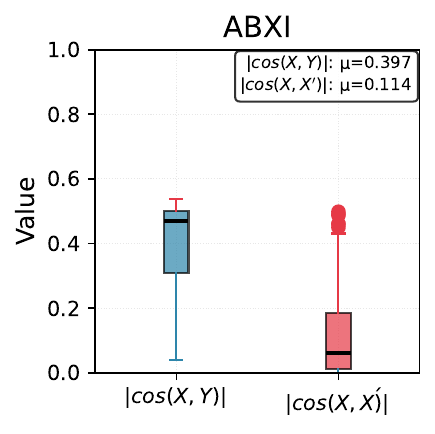}
        \caption{ABXI}
        \label{fig:cos_abxi}
    \end{subfigure}
    \begin{subfigure}[t]{0.27\textwidth}
        \centering
        \includegraphics[width=\textwidth]{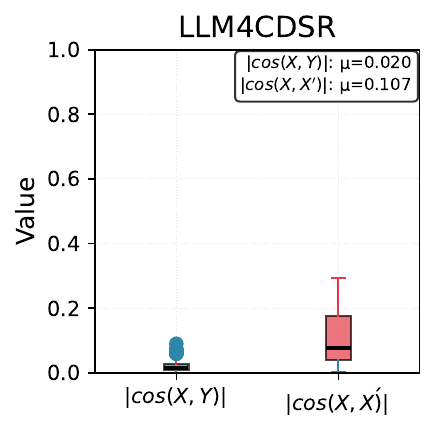}
        \caption{LLM4CDSR}
        \label{fig:cos_llm4cdsr}
    \end{subfigure}
    \caption{Boxplots of cosine similarities $|\cos(X,Y)|$ and $|\cos(X,X')|$. While $|\cos(X,X')|$ remains stable across models (median $\approx \in [0.1,0.2]$), $|\cos(X,Y)|$ varies substantially depending on the dataset, highlighting that $\gca$ induces a consistent degree of orthogonalization regardless of underlying X(query)–Y(key,value) similarity. $\mu$ represents a median.}
    \label{fig:cosXY_cosXX'}
\end{figure}

\begin{itemize}
    \item \textbf{Stability of \boldsymbol{$|\cos(X, X')|$}}. Across all models, the interquartile range of $|\cos(X, X')|$ remains confined to $[0,0.3]$, with medians between $0.1$ and $0.2$, regardless of the dataset pair. For example, in LLM4CDSR the mean $|\cos(X, X')|$ is $0.107$, in CDSRNP it is $0.167$, and in ABXI it is $0.114$. This suggests that $\gca$ consistently enforces a bounded degree of orthogonality between $X$ and $X'$, acting as a regularization mechanism that stabilizes representations.
    \item \textbf{Variability of \boldsymbol{$|\cos(X, Y)|$}}. In contrast, $|\cos(X, Y)|$ shows substantial dataset-dependent variation: very low in LLM4CDSR (mean $0.020$), moderate in CDSRNP ($0.289$), and relatively high in ABXI ($0.397$) . This highlights that query–key alignment is primarily determined by the inherent similarity structure of the source domains, rather than by the $\gca$ module itself.
\end{itemize}

Altogether we conclude that $\gca$ decouples $X'$ from $X$ to a controlled extent, maintaining $|\cos(X, X')|$ within a narrow and stable range even when $|\cos(X, Y)|$ varies widely across datasets. In effect, $\gca$ internally induces orthogonalization independently of how similar $X$ and $Y$ are, ensuring that $X'$ does not simply replicate $X$ or the key,value features, but instead extracts complementary information.


Finally, we summarize above three observations as follows. 

\begin{tcolorbox}[colback=metablue!08, title=\textbf{Summary of Observations}] 
\begin{enumerate}
    \item Applying $\gca$ at the early stage consistently improves performance, whereas vertically stacking multiple $\gca$ modules does not necessarily yield additional gains. 
    \item $\gca$ intrinsically induces the Orthogonal Alignment phenomenon; specifically, it tends to drive the input query and its cross-attended update toward orthogonality, a property that correlates with improved performance. 
    \item A plausible explanation for the emergence of Orthogonal Alignment is that it provides a parameter-efficient strategy for scaling models, yielding a superior accuracy–per–parameter trade-off. 
    \item The orthogonalization effect induced by $\gca$ arises independently of the similarity between $X$ and $Y$, underscoring that its regularization role is not contingent on domain-specific alignment. 
\end{enumerate} 
\end{tcolorbox}

\section{Questions \& Answers} 
\label{sec:QnAs}
Finally, we summarize several questions that readers may naturally raise about our experiments. We hope that the following clarifications provide a deeper understanding of our analysis and help situate this paper’s contribution within the broader literature.  

\textbf{Q1. Is the Orthogonal Alignment a bounded phenomenon dependent on the gated cross-attention mechanism?}  

Based on our experiments, we argue that $\gca$ is one plausible algorithmic instantiation that induces Orthogonal Alignment; that is, its architecture enables the extraction of orthogonal information relative to the query, which in turn contributes to positive transfer from the (query, key) interactions. However, our position is not that Orthogonal Alignment is \emph{bounded} to a particular architectural choice such as $\gca$. Rather, our claim is more general: Orthogonal Alignment is a hidden phenomenon in the alignment problem, and this paper aims primarily to uncover this effect and to analyze why it emerges in practice.  

Therefore, while we identify $\gca$ as a compelling and parameter-efficient mechanism for realizing this property, we also view our work as an invitation to further research. Future directions should explore alternative architectural designs that can more effectively harness Orthogonal Alignment, potentially moving beyond gated cross-attention to achieve even greater utilization of this phenomenon.  

\textbf{Q2. In Table \ref{Table1.Experiment results}, why are different subsets of data used for different baselines?}  

The principal reason is that our work builds directly on publicly available CDSR baselines. For each baseline, we preserve the dataset and experimental configuration originally reported by its authors, and then extend the model with an additional gated cross-attention module to conduct our experiments. This design choice ensures reproducibility: other researchers can readily verify our results by reusing the \emph{exact same baseline implementations} and the \emph{exact same data} under their canonical settings. It is also worth noting that although all three baselines are derived from Amazon user–item interaction logs, each employs its own preprocessing pipeline.    

It is therefore important to emphasize that our findings are not the product of selective reporting or cherry-picking. We did not omit or substitute datasets that might contradict our claims; rather, we intentionally adhered to the established evaluation protocols of the baseline models. This approach strengthens the external validity of our study while maintaining transparency in how results were obtained.  

\textbf{Q3. Does this paper claim that orthogonal alignment is more important than residual alignment?}

No. This paper does not compare the relative importance of orthogonal alignment and residual alignment. Instead, we claim that both mechanisms can occur simultaneously within the cross-attention algorithm to enhance performance. We clearly acknowledge prior literature demonstrating the role of cross-attention in residual alignment, but we did not explicitly investigate residual alignment in our experiments. The main contribution of this work is to uncover a new mechanism—orthogonal alignment—alongside the previously recognized residual alignment.

\textbf{Q4. Is this work expandable to vision-language models?}

We are cautious about generalizing our findings to vision-language models, as our experiments on orthogonal alignment were conducted exclusively with recommendation data. However, we expect optimistically that the orthogonal alignment phenomenon could also be observed in vision-language models. Our study used transformer-based architectures with gated cross-attention, which are commonly employed in vision-language research.
The main differences between our setting and typical vision-language models are:
\begin{enumerate}
    \item We observed orthogonal alignment using recommendation data.
    \item Vision-language models typically use pre-trained image and text encoders as raw representations.
\end{enumerate}
It may be more challenging to observe orthogonal alignment in vision-language models because their encoders are often trained with self-contrastive learning, which encourages high cosine similarity for matching image-text pairs and low similarity for non-matching pairs. As a result, the latent vectors for text and image pairs are already well aligned before cross-attention is applied. In contrast, our experiments did not involve self-contrastive pre-training for the recommendation data encoders.
Therefore, while we expect orthogonal alignment to occur in vision-language models, it may arise under more nuanced conditions due to the pre-aligned nature of their representations.

\section{Conclusion}

This work fundamentally reconceptualizes cross-domain alignment mechanisms in sequential recommendation systems, challenging the prevailing orthodoxy that effective cross-modal fusion necessitates parallelism-based approaches. Our investigation reveals that \textbf{Orthogonal Alignment}—wherein cross-attention mechanisms spontaneously extract orthogonal, complementary information—constitutes a more efficacious paradigm for positive transfer learning. Our central contribution lies not merely in identifying this orthogonal phenomenon, but in elucidating its mechanistic origins as a parameter-efficient scaling strategy that exploits previously underutilized representational subspaces.

The implications of these findings extend far beyond the immediate domain of cross-domain sequential recommendation, suggesting fundamental reconsiderations across the broader landscape of multi-modal representation learning. The transition toward orthogonality-aware multi-modal learning represents a paradigmatic shift that promises to unlock previously inaccessible representational capacity while maintaining parameter efficiency—a critical advancement for the scalable deployment of sophisticated multi-modal systems across resource-constrained environments. The development of nuanced orthogonal alignment measures constitutes not merely a methodological improvement, but a fundamental reconceptualization of how we understand and optimize cross-modal information fusion in the era of large-scale representation learning.

\section{Acknowledgment}

We would like to first thank Rui Li, Jawed Bappy, Ming Li, Will Fu, Mingfu Liang, and Michael Louis luzzolino from the Ranking AI Research team at Meta for their valuable discussions during the early stages of this work. We are also grateful to the Meta leadership for their support of computational resources, as well as for organizing open discussion sessions to share this work and gather diverse feedback within the Ranking AI Research team. Additionally, we thank Srikar Babu, Yixiao Hwang, Seohong Park from UC Berkeley, and Chanwoo Park from MIT for taking the time to review an early draft and provide helpful feedback.

\clearpage
\newpage
\bibliographystyle{assets/plainnat}
\bibliography{paper}

\appendix
\section{Experiment results}
We now provide further details on the experimental results reported in Table~\ref{Table1.Experiment results}. In the model names, the terms \texttt{sigmoid} and \texttt{tanh} indicate the activation function used in the gating module, while \texttt{true} and \texttt{false} specify whether a \texttt{LayerNorm} operator is included. For example, the notation $+\gca[1]\text{-sigmoid, True}$ denotes that a $\gca$ module is inserted at position $1$, the gating module employs a Sigmoid activation function, and a \texttt{LayerNorm} operator is applied.
\subsection{ABXI}
\begin{table}[H]
\centering
\small
\setlength{\tabcolsep}{2pt}
\resizebox{\textwidth}{!}{%
\begin{tabular}{llHlHHHHHlHlHHHHHll}
\toprule

model & NDCG@1 A & NDCG@5 A & NDCG@10 A & NDCG@20 A & HR@1 A & HR@5 A & HR@10 A & HR@20 A & NDCG@1 B & NDCG@5 B & NDCG@10 B & NDCG@20 B & HR@1 B & HR@5 B & HR@10 B & HR@20 B & AUC@A & AUC@B \\ \midrule

ABXI & 0.0593 \tiny{$\pm$ 0.0074} & 0.1279 $\pm$ 0.0140 & 0.1541 \tiny{$\pm$ 0.0130} & 0.1701 $\pm$ 0.0103 & 0.0593 $\pm$ 0.0074 & 0.1959 $\pm$ 0.0199 & 0.2766 $\pm$ 0.0166 & 0.3394 $\pm$ 0.0063 & 0.0416  \tiny{$\pm$ 0.0058} & 0.0870 $\pm$ 0.0100 & 0.1093  \tiny{$\pm$ 0.0113} & 0.1316 $\pm$ 0.0113 & 0.0416 $\pm$ 0.0058 & 0.1316 $\pm$ 0.0148 & 0.2010 $\pm$ 0.0189 & 0.2893 $\pm$ 0.0186 & 0.7205  \tiny{$\pm$ 0.0015} & 0.7180  \tiny{$\pm$ 0.0032} \\

$+\gca[0]$-sigmoid,False & 0.0724 \tiny{$\pm$ 0.0094} & 0.1509 $\pm$ 0.0098 & 0.1763 \tiny{$\pm$ 0.0092} & 0.1892 $\pm$ 0.0083 & 0.0724 $\pm$ 0.0094 & 0.2277 $\pm$ 0.0109 & 0.3053 $\pm$ 0.0085 & 0.3566 $\pm$ 0.0054 & 0.0525  \tiny{$\pm$ 0.0053} & 0.1050  $\pm$ 0.0068 & 0.1289  \tiny{$\pm$ 0.0072} & 0.1486 $\pm$ 0.0051 & 0.0525 $\pm$ 0.0053 & 0.1562 $\pm$ 0.0080 & 0.2304 $\pm$ 0.0092 & 0.3084 $\pm$ 0.0028 & 0.7305  \tiny{$\pm$ 0.0039} & 0.7204  \tiny{$\pm$ 0.0031} \\

$+\gca[1]$-sigmoid,False  & 0.0715 \tiny{$\pm$ 0.0068} & 0.1479 $\pm$ 0.0058 & 0.1720 \tiny{$\pm$ 0.0046} & 0.1853 $\pm$ 0.0039 & 0.0715 $\pm$ 0.0068 & 0.2218 $\pm$ 0.0056 & 0.2957 $\pm$ 0.0038 & 0.3483 $\pm$ 0.0021 & 0.0489  \tiny{$\pm$ 0.0034} & 0.0994 $\pm$ 0.0068 & 0.1223  \tiny{$\pm$ 0.0066} & 0.1424 $\pm$ 0.0048 & 0.0489 $\pm$ 0.0034 & 0.1490 $\pm$ 0.0096 & 0.2205 $\pm$ 0.0094 & 0.2996 $\pm$ 0.0023 & 0.7285  \tiny{$\pm$ 0.0022} & 0.7047  \tiny{$\pm$ 0.0099} \\

$+\gca[0,1]$-sigmoid,False & 0.0654 \tiny{$\pm$ 0.0034} & 0.1422 $\pm$ 0.0032 & 0.1674 \tiny{$\pm$ 0.0027} & 0.1815 $\pm$ 0.0024 & 0.0654 $\pm$ 0.0034 & 0.2173 $\pm$ 0.0053 & 0.2946 $\pm$ 0.0058 & 0.3506 $\pm$ 0.0033 & 0.0458  \tiny{$\pm$ 0.0019} & 0.0942 $\pm$ 0.0010 & 0.1170  \tiny{$\pm$ 0.0010} & 0.1385 $\pm$ 0.0012 & 0.0458 $\pm$ 0.0019 & 0.1425 $\pm$ 0.0014 & 0.2133 $\pm$ 0.0036 & 0.2982 $\pm$ 0.0041 & 0.7290  \tiny{$\pm$ 0.0017} & 0.7116  \tiny{$\pm$ 0.0015} \\

$+\gca[0,2]$-sigmoid,False & 0.0735 \tiny{$\pm$ 0.0073} & 0.1456 $\pm$ 0.0061 & 0.1697 \tiny{$\pm$ 0.0042} & 0.1844 $\pm$ 0.0045 & 0.0735 $\pm$ 0.0073 & 0.2169 $\pm$ 0.0061 & 0.2914 $\pm$ 0.0050 & 0.3492 $\pm$ 0.0048 & 0.0490  \tiny{$\pm$ 0.0026} & 0.0971 $\pm$ 0.0050 & 0.1197  \tiny{$\pm$ 0.0046} & 0.1399 $\pm$ 0.0038 & 0.0490 $\pm$ 0.0026 & 0.1446 $\pm$ 0.0073 & 0.2147 $\pm$ 0.0061 & 0.2948 $\pm$ 0.0033 & 0.7282  \tiny{$\pm$ 0.0014} & 0.7075  \tiny{$\pm$ 0.0048} \\

$+\gca[1,2]$-sigmoid,False & 0.0732 \tiny{$\pm$ 0.0078} & 0.1460 $\pm$ 0.0073 & 0.1699 \tiny{$\pm$ 0.0072} & 0.1832 $\pm$ 0.0064 & 0.0732 $\pm$ 0.0078 & 0.2161 $\pm$ 0.0069 & 0.2898 $\pm$ 0.0068 & 0.3422 $\pm$ 0.0039 & 0.0505  \tiny{$\pm$ 0.0056} & 0.1001 $\pm$ 0.0067 & 0.1222  \tiny{$\pm$ 0.0058} & 0.1432 $\pm$ 0.0050 & 0.0505 $\pm$ 0.0056 & 0.1490 $\pm$ 0.0077 & 0.2178 $\pm$ 0.0047 & 0.3006 $\pm$ 0.0020 & 0.7188  \tiny{$\pm$ 0.0051} & 0.6998  \tiny{$\pm$ 0.0041} \\

$+\gca[0,1,2]$-sigmoid,False & 0.0710 \tiny{$\pm$ 0.0037} & 0.1417 $\pm$ 0.0035 & 0.1656 \tiny{$\pm$ 0.0027} & 0.1802 $\pm$ 0.0019 & 0.0710 $\pm$ 0.0037 & 0.2105 $\pm$ 0.0051 & 0.2837 $\pm$ 0.0017 & 0.3415 $\pm$ 0.0023 & 0.0482  \tiny{$\pm$ 0.0012} & 0.0965 $\pm$ 0.0023 & 0.1189  \tiny{$\pm$ 0.0023} & 0.1404 $\pm$ 0.0020 & 0.0482 $\pm$ 0.0012 & 0.1439 $\pm$ 0.0041 & 0.2132 $\pm$ 0.0045 & 0.2980 $\pm$ 0.0041 & 0.7223  \tiny{$\pm$ 0.0008} & 0.7045  \tiny{$\pm$ 0.0032} \\

$+\gca[0]$-tanh,False& 0.0686 \tiny{$\pm$ 0.0031} & 0.1471 $\pm$ 0.0058 & 0.1722 \tiny{$\pm$ 0.0040} & 0.1856 $\pm$ 0.0037 & 0.0686 $\pm$ 0.0031 & 0.2228 $\pm$ 0.0089 & 0.2997 $\pm$ 0.0050 & 0.3523 $\pm$ 0.0042 & 0.0490  \tiny{$\pm$ 0.0054} & 0.1002 $\pm$ 0.0083 & 0.1244  \tiny{$\pm$ 0.0071} & 0.1457 $\pm$ 0.0049 & 0.0490 $\pm$ 0.0054 & 0.1501 $\pm$ 0.0108 & 0.2257 $\pm$ 0.0072 & 0.3095 $\pm$ 0.0045 & 0.7273  \tiny{$\pm$ 0.0042} & 0.7210  \tiny{$\pm$ 0.0015} \\

$+\gca[1]$-tanh,False & 0.0727 \tiny{$\pm$ 0.0098} & 0.1483 $\pm$ 0.0090 & 0.1720  \tiny{$\pm$ 0.0072} & 0.1859 $\pm$ 0.0067 & 0.0727 $\pm$ 0.0098 & 0.2215 $\pm$ 0.0077 & 0.2935 $\pm$ 0.0021 & 0.3483 $\pm$ 0.0011 & 0.0461  \tiny{$\pm$ 0.0035} & 0.0972 $\pm$ 0.0051 & 0.1200  \tiny{$\pm$ 0.0044} & 0.1409 $\pm$ 0.0027 & 0.0461 $\pm$ 0.0035 & 0.1473 $\pm$ 0.0073 & 0.2177 $\pm$ 0.0050 & 0.3006 $\pm$ 0.0028 & 0.7281  \tiny{$\pm$ 0.0011} & 0.7063  \tiny{$\pm$ 0.0093} \\

$+\gca[0,1]$-tanh,False& 0.0677 \tiny{$\pm$ 0.0041} & 0.1439 $\pm$ 0.0027 & 0.1697  \tiny{$\pm$ 0.0018} & 0.1832 $\pm$ 0.0025 & 0.0677 $\pm$ 0.0041 & 0.2194 $\pm$ 0.0032 & 0.2988 $\pm$ 0.0025 & 0.3517 $\pm$ 0.0023 & 0.0479 \tiny{$\pm$ 0.0018} & 0.0969 $\pm$ 0.0026 & 0.1201  \tiny{$\pm$ 0.0024} & 0.1414 $\pm$ 0.0019 & 0.0479 $\pm$ 0.0018 & 0.1454 $\pm$ 0.0042 & 0.2176 $\pm$ 0.0034 & 0.3019 $\pm$ 0.0017 & 0.7302  \tiny{$\pm$ 0.0014} & 0.7125  \tiny{$\pm$ 0.0031} \\

$+\gca[0,2]$-tanh,False & 0.0679 \tiny{$\pm$ 0.0047} & 0.1405 $\pm$ 0.0059 & 0.1650  \tiny{$\pm$ 0.0040} & 0.1796 $\pm$ 0.0040 & 0.0679 $\pm$ 0.0047 & 0.2111 $\pm$ 0.0073 & 0.2866 $\pm$ 0.0016 & 0.3444 $\pm$ 0.0028 & 0.0489 \tiny{$\pm$ 0.0029} & 0.0983 $\pm$ 0.0028 & 0.1217  \tiny{$\pm$ 0.0023} & 0.1420 $\pm$ 0.0017 & 0.0489 $\pm$ 0.0029 & 0.1459 $\pm$ 0.0047 & 0.2187 $\pm$ 0.0038 & 0.2987 $\pm$ 0.0013 & 0.7265  \tiny{$\pm$ 0.0035} & 0.7051  \tiny{$\pm$ 0.0051} \\

$+\gca[0,1,2]$-tanh,False& 0.0722 \tiny{$\pm$ 0.0078} & 0.1469 $\pm$ 0.0067 & 0.1692  \tiny{$\pm$ 0.0061} & 0.1834 $\pm$ 0.0056 & 0.0722 $\pm$ 0.0078 & 0.2203 $\pm$ 0.0053 & 0.2888 $\pm$ 0.0044 & 0.3448 $\pm$ 0.0025 & 0.0490  \tiny{$\pm$ 0.0031} & 0.0965 $\pm$ 0.0040 & 0.1198  \tiny{$\pm$ 0.0036} & 0.1404 $\pm$ 0.0023 & 0.0490 $\pm$ 0.0031 & 0.1431 $\pm$ 0.0045 & 0.2155 $\pm$ 0.0032 & 0.2967 $\pm$ 0.0040 & 0.7197  \tiny{$\pm$ 0.0066} & 0.7013  \tiny{$\pm$ 0.0046} \\

$+\gca[1,2]$-tanh,False& 0.0770 \tiny{$\pm$ 0.0061} & 0.1528 $\pm$ 0.0072 & 0.1761  \tiny{$\pm$ 0.0051} & 0.1890 $\pm$ 0.0056 & 0.0770 $\pm$ 0.0061 & 0.2253 $\pm$ 0.0078 & 0.2968 $\pm$ 0.0019 & 0.3481 $\pm$ 0.0058 & 0.0476  \tiny{$\pm$ 0.0049} & 0.0984 $\pm$ 0.0082 & 0.1206  \tiny{$\pm$ 0.0079} & 0.1400 $\pm$ 0.0065 & 0.0476 $\pm$ 0.0049 & 0.1480 $\pm$ 0.0105 & 0.2171 $\pm$ 0.0098 & 0.2936 $\pm$ 0.0043 & 0.7222  \tiny{$\pm$ 0.0080} & 0.6988  \tiny{$\pm$ 0.0080} \\

$+\gca[0]$-sigmoid,True & 0.0703 \tiny{$\pm$ 0.0000} & 0.1531 $\pm$ 0.0000 & 0.1757  \tiny{$\pm$ 0.0000} & 0.1907 $\pm$ 0.0000 & 0.0703 $\pm$ 0.0000 & 0.2336 $\pm$ 0.0000 & 0.3022 $\pm$ 0.0000 & 0.3605 $\pm$ 0.0000 & 0.0548  \tiny{$\pm$ 0.0000} & 0.1096 $\pm$ 0.0000 & 0.1327  \tiny{$\pm$ 0.0000} & 0.1497 $\pm$ 0.0000 & 0.0548 $\pm$ 0.0000 & 0.1621 $\pm$ 0.0000 & 0.2341 $\pm$ 0.0000 & 0.3010 $\pm$ 0.0000 & 0.7317  \tiny{$\pm$ 0.0000} & 0.7150  \tiny{$\pm$ 0.0000} \\

$+\gca[0,1]$-sigmoid,True & 0.0678 \tiny{$\pm$ 0.0000} & 0.1447 $\pm$ 0.0000 & 0.1695  \tiny{$\pm$ 0.0000} & 0.1828 $\pm$ 0.0000 & 0.0678 $\pm$ 0.0000 & 0.2203 $\pm$ 0.0000 & 0.2968 $\pm$ 0.0000 & 0.3489 $\pm$ 0.0000 & 0.0485  \tiny{$\pm$ 0.0000} & 0.0946 $\pm$ 0.0000 & 0.1184  \tiny{$\pm$ 0.0000} & 0.1379 $\pm$ 0.0000 & 0.0485 $\pm$ 0.0000 & 0.1401 $\pm$ 0.0000 & 0.2135 $\pm$ 0.0000 & 0.2904 $\pm$ 0.0000 & 0.7257  \tiny{$\pm$ 0.0000} & 0.6962  \tiny{$\pm$ 0.0000} \\

$+\gca[0,2]$-sigmoid,True & 0.0807 \tiny{$\pm$ 0.0033} & 0.1514 $\pm$ 0.0032 & 0.1732  \tiny{$\pm$ 0.0027} & 0.1881 $\pm$ 0.0028 & 0.0807 $\pm$ 0.0033 & 0.2192 $\pm$ 0.0032 & 0.2861 $\pm$ 0.0022 & 0.3450 $\pm$ 0.0038 & 0.0516  \tiny{$\pm$ 0.0024} & 0.0991 $\pm$ 0.0019 & 0.1211  \tiny{$\pm$ 0.0020} & 0.1384 $\pm$ 0.0023 & 0.0516 $\pm$ 0.0024 & 0.1453 $\pm$ 0.0037 & 0.2136 $\pm$ 0.0041 & 0.2817 $\pm$ 0.0074 & 0.7235  \tiny{$\pm$ 0.0038} & 0.6996  \tiny{$\pm$ 0.0028} \\

$+\gca[1]$-sigmoid,True& 0.0815 \tiny{$\pm$ 0.0037} & 0.1553 $\pm$ 0.0035 & 0.1767  \tiny{$\pm$ 0.0041} & 0.1897 $\pm$ 0.0036 & 0.0815 $\pm$ 0.0037 & 0.2264 $\pm$ 0.0038 & 0.2924 $\pm$ 0.0061 & 0.3437 $\pm$ 0.0053 & 0.0559  \tiny{$\pm$ 0.0019} & 0.1086 $\pm$ 0.0024 & 0.1315  \tiny{$\pm$ 0.0022} & 0.1490 $\pm$ 0.0018 & 0.0559 $\pm$ 0.0019 & 0.1603 $\pm$ 0.0032 & 0.2311 $\pm$ 0.0051 & 0.2999 $\pm$ 0.0035 & 0.7218  \tiny{$\pm$ 0.0061} & 0.7023  \tiny{$\pm$ 0.0019} \\

$+\gca[1,2]$-sigmoid,True & 0.0904 \tiny{$\pm$ 0.0027} & 0.1612 $\pm$ 0.0031 & 0.1831  \tiny{$\pm$ 0.0023} & 0.1964 $\pm$ 0.0023 & 0.0904 $\pm$ 0.0027 & 0.2288 $\pm$ 0.0058 & 0.2963 $\pm$ 0.0037 & 0.3488 $\pm$ 0.0054 & 0.0511  \tiny{$\pm$ 0.0039} & 0.1000 $\pm$ 0.0046 & 0.1217  \tiny{$\pm$ 0.0049} & 0.1393 $\pm$ 0.0045 & 0.0511 $\pm$ 0.0039 & 0.1483 $\pm$ 0.0056 & 0.2156 $\pm$ 0.0070 & 0.2850 $\pm$ 0.0058 & 0.7180  \tiny{$\pm$ 0.0071} & 0.6846  \tiny{$\pm$ 0.0023} \\

$+\gca[0,1,2]$-sigmoid,True & 0.0690 \tiny{$\pm$ 0.0000} & 0.1435 $\pm$ 0.0000 & 0.1683  \tiny{$\pm$ 0.0000} & 0.1839 $\pm$ 0.0000 & 0.0690 $\pm$ 0.0000 & 0.2154 $\pm$ 0.0000 & 0.2914 $\pm$ 0.0000 & 0.3530 $\pm$ 0.0000 & 0.0497  \tiny{$\pm$ 0.0000} & 0.0963 $\pm$ 0.0000 & 0.1189  \tiny{$\pm$ 0.0000} & 0.1397 $\pm$ 0.0000 & 0.0497 $\pm$ 0.0000 & 0.1426 $\pm$ 0.0000 & 0.2129 $\pm$ 0.0000 & 0.2952 $\pm$ 0.0000 & 0.7303  \tiny{$\pm$ 0.0000} & 0.6937  \tiny{$\pm$ 0.0000} \\

$+\gca[1]$-tanh,True & 0.0808 \tiny{$\pm$ 0.0061} & 0.1537 $\pm$ 0.0066 & 0.1749  \tiny{$\pm$ 0.0058} & 0.1875 $\pm$ 0.0054 & 0.0808 $\pm$ 0.0061 & 0.2236 $\pm$ 0.0077 & 0.2883 $\pm$ 0.0045 & 0.3385 $\pm$ 0.0039 & 0.0542  \tiny{$\pm$ 0.0037} & 0.1069 $\pm$ 0.0038 & 0.1302  \tiny{$\pm$ 0.0033} & 0.1467 $\pm$ 0.0035 & 0.0542 $\pm$ 0.0037 & 0.1584 $\pm$ 0.0038 & 0.2305 $\pm$ 0.0026 & 0.2953 $\pm$ 0.0026 & 0.7138  \tiny{$\pm$ 0.0036} & 0.6917  \tiny{$\pm$ 0.0021} \\

$+\gca[1,2]$-tanh,True & 0.0882 \tiny{$\pm$ 0.0052} & 0.1649 $\pm$ 0.0024 & 0.1853  \tiny{$\pm$ 0.0013} & 0.1978 $\pm$ 0.0019 & 0.0882 $\pm$ 0.0052 & 0.2377 $\pm$ 0.0075 & 0.3001 $\pm$ 0.0033 & 0.3494 $\pm$ 0.0058 & 0.0527  \tiny{$\pm$ 0.0020} & 0.1055 $\pm$ 0.0031 & 0.1282  \tiny{$\pm$ 0.0028} & 0.1461 $\pm$ 0.0025 & 0.0527 $\pm$ 0.0020 & 0.1560 $\pm$ 0.0049 & 0.2265 $\pm$ 0.0041 & 0.2972 $\pm$ 0.0030 & 0.7148  \tiny{$\pm$ 0.0026} & 0.6924  \tiny{$\pm$ 0.0009} \\
\bottomrule
\end{tabular}
}
\caption{ABXI and ABXI+$\gca$ on Amazon Food-Kitchen dataset.}
\label{tab:afk_metrics_extracted}
\end{table}

\begin{table}[H]
\centering
\small
\setlength{\tabcolsep}{2pt}
\resizebox{\textwidth}{!}{%
\begin{tabular}{llHlHHHHHlHlHHHHHll}
\toprule
model & NDCG@1 A & NDCG@5 A & NDCG@10 A & NDCG@20 A & HR@1 A & HR@5 A & HR@10 A & HR@20 A & NDCG@1 B & NDCG@5 B & NDCG@10 B & NDCG@20 B & HR@1 B & HR@5 B & HR@10 B & HR@20 B & AUC@A & AUC@B \\ \midrule

ABXI & 0.0730 {\tiny $\pm$0.0070} & 0.1458$\pm$0.0093 & 0.1724 {\tiny$\pm$0.0071} & 0.2041$\pm$0.0060 & 0.0730$\pm$0.0070 & 0.2183$\pm$0.0122 & 0.3318$\pm$0.0055 & 0.4170$\pm$0.0025 & 0.0548  \tiny{$\pm$0.0038} & 0.1071$\pm$0.0033 & 0.1273 \tiny{$\pm$0.0028} & 0.1482$\pm$0.0028 & 0.0548$\pm$0.0038 & 0.1567$\pm$0.0031 & 0.2191$\pm$0.0030 & 0.3022$\pm$0.0053 & 0.7216 \tiny{$\pm$0.0027} & 0.7123 \tiny{$\pm$0.0009} \\

$+\gca[1]$-sigmoid, True & 0.0711 {\tiny $\pm$0.0057} & 0.1375 $\pm$0.0094 & 0.1747 {\tiny$\pm$0.0069} & 0.1989$\pm$0.0050 & 0.0711$\pm$0.0057 & 0.2042$\pm$0.0140 & 0.3196$\pm$0.0068 & 0.4141$\pm$0.0028 & 0.0560  \tiny{$\pm$0.0020} & 0.1056$\pm$0.0031 & 0.1283 \tiny{$\pm$0.0031} & 0.1486$\pm$0.0020 & 0.0560$\pm$0.0020 & 0.1534$\pm$0.0055 & 0.2238$\pm$0.0062 & 0.3043$\pm$0.0052 & 0.7409 \tiny{$\pm$0.0019} & 0.7057 \tiny{$\pm$0.0040} \\

$+\gca[1,2]$-tanh, True & 0.0733 {\tiny $\pm$0.0042} & 0.1466$\pm$0.0056 & 0.1846 {\tiny$\pm$0.0057} & 0.2079$\pm$0.0042 & 0.0733$\pm$0.0042 & 0.2200$\pm$0.0071 & 0.3377$\pm$0.0080 & 0.4292$\pm$0.0044 & 0.0566 \tiny{$\pm$0.0052} & 0.1052$\pm$0.0041 & 0.1271 \tiny{$\pm$0.0042} & 0.1494$\pm$0.0049 & 0.0566$\pm$0.0052 & 0.1517$\pm$0.0044 & 0.2198$\pm$0.0041 & 0.3080$\pm$0.0078 & 0.7354 \tiny{$\pm$0.0048} & 0.6973 \tiny{$\pm$0.0051} \\

$+\gca[0]$-sigmoid, True & 0.0727 {\tiny $\pm$0.0060} & 0.1416$\pm$0.0054 & 0.1793 {\tiny$\pm$0.0047} & 0.2039$\pm$0.0040 & 0.0727$\pm$0.0060 & 0.2103$\pm$0.0061 & 0.3273$\pm$0.0038 & 0.4231$\pm$0.0029 & 0.0544 \tiny{$\pm$0.0044} & 0.1034$\pm$0.0030 & 0.1244 \tiny{$\pm$0.0025} & 0.1446$\pm$0.0033 & 0.0544$\pm$0.0044 & 0.1501$\pm$0.0022 & 0.2152$\pm$0.0015 & 0.2954$\pm$0.0040 & 0.7410 \tiny{$\pm$0.0025} & 0.7169 \tiny{$\pm$0.0024} \\

$+\gca[0,2]$-sigmoid, True & 0.0707 {\tiny $\pm$0.0038} & 0.1335$\pm$0.0043 & 0.1654 {\tiny$\pm$0.0029} & 0.1865$\pm$0.0026 & 0.0707$\pm$0.0038 & 0.1985$\pm$0.0046 & 0.2969$\pm$0.0019 & 0.3797$\pm$0.0032 & 0.0511 \tiny{$\pm$0.0042} & 0.0952$\pm$0.0059 & 0.1158 \tiny{$\pm$0.0058} & 0.1365$\pm$0.0054 & 0.0511$\pm$0.0042 & 0.1376$\pm$0.0078 & 0.2016$\pm$0.0084 & 0.2835$\pm$0.0071 & 0.7397 \tiny{$\pm$0.0023} & 0.7089 \tiny{$\pm$0.0035} \\

$+\gca[1]$-tanh, True & 0.0727 {\tiny$\pm$0.0045} & 0.1398$\pm$0.0078 & 0.1769 {\tiny $\pm$0.0059} & 0.2015$\pm$0.0050 & 0.0727$\pm$0.0045 & 0.2092$\pm$0.0122 & 0.3243$\pm$0.0063 & 0.4205$\pm$0.0036 & 0.0531 \tiny{$\pm$0.0014} & 0.1023$\pm$0.0036 & 0.1233 \tiny{$\pm$0.0024} & 0.1441$\pm$0.0017 & 0.0531$\pm$0.0014 & 0.1503$\pm$0.0063 & 0.2155$\pm$0.0031 & 0.2982$\pm$0.0022 & 0.7370 \tiny{$\pm$0.0014} & 0.6994 \tiny{$\pm$0.0051} \\

$+\gca[1,2]$-sigmoid, True & 0.0701 {\tiny$\pm$0.0025} & 0.1355$\pm$0.0019 & 0.1706 {\tiny$\pm$0.0029} & 0.1939$\pm$0.0032 & 0.0701$\pm$0.0025 & 0.2008$\pm$0.0033 & 0.3093$\pm$0.0073 & 0.4017$\pm$0.0083 & 0.0513 \tiny{$\pm$0.0027} & 0.0955$\pm$0.0035 & 0.1163 \tiny{$\pm$0.0031} & 0.1365$\pm$0.0022 & 0.0513$\pm$0.0027 & 0.1376$\pm$0.0057 & 0.2022$\pm$0.0048 & 0.2825$\pm$0.0036 & 0.7359 \tiny{$\pm$0.0010} & 0.7046 \tiny{$\pm$0.0042} \\

$+\gca[0]$-tanh, True & 0.0731 {\tiny$\pm$0.0045} & 0.1390$\pm$0.0052 & 0.1773 {\tiny$\pm$0.0064} & 0.2030$\pm$0.0045 & 0.0731$\pm$0.0045 & 0.2051$\pm$0.0070 & 0.3240$\pm$0.0088 & 0.4249$\pm$0.0017 & 0.0551 \tiny{$\pm$0.0047} & 0.1025$\pm$0.0025 & 0.1245 \tiny{$\pm$0.0017} & 0.1450$\pm$0.0025 & 0.0551$\pm$0.0047 & 0.1481$\pm$0.0020 & 0.2163$\pm$0.0021 & 0.2974$\pm$0.0019 & 0.7399 \tiny{$\pm$0.0022} & 0.7172 \tiny{$\pm$0.0027} \\

$+\gca[0,2]$-tanh, True & 0.0693 {\tiny$\pm$0.0018} & 0.1316$\pm$0.0037 & 0.1634 {\tiny$\pm$0.0027} & 0.1842$\pm$0.0032 & 0.0693$\pm$0.0018 & 0.1951$\pm$0.0061 & 0.2928$\pm$0.0040 & 0.3744$\pm$0.0051 & 0.0528 \tiny{$\pm$0.0026} & 0.0962$\pm$0.0042 & 0.1164 \tiny{$\pm$0.0039} & 0.1368$\pm$0.0031 & 0.0528$\pm$0.0026 & 0.1380$\pm$0.0072 & 0.2009$\pm$0.0054 & 0.2819$\pm$0.0023 & 0.7374 \tiny{$\pm$0.0014} & 0.7118 \tiny{$\pm$0.0037} \\
\bottomrule
\end{tabular}
}
\caption{ABXI and ABXI+$\gca$ on Amazon Beauty-Electronics dataset}
\label{tab:metrics_extracted}
\end{table}

\subsection{LLM4CDSR}
For experiments where \texttt{LLM4CDSR} serves as the baseline, we fix the LayerNorm operator to \texttt{True} and vary both the activation function in the gating module (Sigmoid or Tanh) and the number of cross-attention heads (4 or 8). For example, the notation $+\gca[0,1]\text{-Sigmoid,4}$ denotes that two $\gca$ modules are inserted at positions $0$ and $1$, where the gating function uses a Sigmoid activation and the cross-attention mechanism employs $4$ heads in total.
\begin{table}[H]
\centering
\small
\setlength{\tabcolsep}{2pt}
\resizebox{\textwidth}{!}{%
\begin{tabular}{lcHcHHHHHcHcHHHHHcc}
\toprule
Model & NDCG@1 A & NDCG@5 A & NDCG@10 A & NDCG@20 A & HR@1 & HR@5 A & HR@10 A & HR@20 A & NDCG@1 B & NDCG@5 B & NDCG@10 B & NDCG@20 B & HR@1 B & HR@5 B & HR@10 B & HR@20 B & AUC A & AUC B \\ \midrule

\texttt{LLLM4CDSR} & 0.7157 \tiny{$\pm$ 0.0025} & 0.7836 $\pm$ 0.0015 & 0.7821 \tiny{$\pm$ 0.0018} & 0.8050 $\pm$ 0.0016 & 0.7157 $\pm$ 0.0025 & 0.8172 $\pm$ 0.0015 & 0.8526 $\pm$ 0.0030 & 0.8919 $\pm$ 0.0031 & 0.5870 \tiny{$\pm$ 0.0051} & 0.6338 $\pm$ 0.0037 & 0.6493 \tiny{$\pm$ 0.0026} & 0.6640 $\pm$ 0.0033 & 0.5870 $\pm$ 0.0051 & 0.6761 $\pm$ 0.0046 & 0.7244 $\pm$ 0.0033 & 0.7824 $\pm$ 0.0066 & 0.9216 \tiny{$\pm$ 0.0013} & 0.8621 \tiny{$\pm$ 0.0054} \\

+$\gca$[0,1]-sigmoid,4 & 0.7310 \tiny{$\pm$ 0.0012} & 0.7949 $\pm$ 0.0012 & 0.8056 \tiny{$\pm$ 0.0014} & 0.8147 $\pm$ 0.0010 & 0.7310 $\pm$ 0.0012 & 0.8267 $\pm$ 0.0014 & 0.8598 $\pm$ 0.0026 & 0.8961 $\pm$ 0.0026 & 0.6112 \tiny{$\pm$ 0.0032} & 0.6507 $\pm$ 0.0029 & 0.6638 \tiny{$\pm$ 0.0038} & 0.6773 $\pm$ 0.0037 & 0.6112 $\pm$ 0.0032 & 0.6863 $\pm$ 0.0034 & 0.7272 $\pm$ 0.0065 & 0.7808 $\pm$ 0.0072 & 0.9370 \tiny{$\pm$ 0.0010} & 0.8664 \tiny{$\pm$ 0.0030} \\

+$\gca$[0,1]-sigmoid,8 & 0.7307 \tiny{$\pm$ 0.0017} & 0.7950 $\pm$ 0.0008 & 0.8056 \tiny{$\pm$ 0.0004} & 0.8146 $\pm$ 0.0003 & 0.7307 $\pm$ 0.0017 & 0.8268 $\pm$ 0.0008 & 0.8596 $\pm$ 0.0014 & 0.8954 $\pm$ 0.0016 & 0.6101 \tiny{$\pm$ 0.0025} & 0.6509 $\pm$ 0.0027 & 0.6644 \tiny{$\pm$ 0.0029} & 0.6772 $\pm$ 0.0023 & 0.6101 $\pm$ 0.0025 & 0.6874 $\pm$ 0.0044 & 0.7294 $\pm$ 0.0075 & 0.7804 $\pm$ 0.0065 & 0.9371 \tiny{$\pm$ 0.0006} & 0.8664 \tiny{$\pm$ 0.0019} \\

+$\gca$[0,1]-tanh,4 & 0.7294 \tiny{$\pm$ 0.0012} & 0.7938 $\pm$ 0.0012 & 0.8044 \tiny{$\pm$ 0.0010} & 0.8131 $\pm$ 0.0005 & 0.7294 $\pm$ 0.0012 & 0.8258 $\pm$ 0.0011 & 0.8590 $\pm$ 0.0008 & 0.8935 $\pm$ 0.0022 & 0.6094 \tiny{$\pm$ 0.0020} & 0.6493 $\pm$ 0.0018 & 0.6632 \tiny{$\pm$ 0.0028} & 0.6756 $\pm$ 0.0023 & 0.6094 $\pm$ 0.0020 & 0.6853 $\pm$ 0.0039 & 0.7284 $\pm$ 0.0068 & 0.7777 $\pm$ 0.0048 & 0.9365 \tiny{$\pm$ 0.0005} & 0.8652 \tiny{$\pm$ 0.0006} \\

+$\gca$[0,1]-tanh,8 & 0.7280 \tiny{$\pm$ 0.0043} & 0.7927 $\pm$ 0.0034 & 0.8032 \tiny{$\pm$ 0.0030} & 0.8125 $\pm$ 0.0030 & 0.7280 $\pm$ 0.0043 & 0.8251 $\pm$ 0.0027 & 0.8576 $\pm$ 0.0021 & 0.8943 $\pm$ 0.0020 & 0.6079 \tiny{$\pm$ 0.0036} & 0.6485 $\pm$ 0.0035 & 0.6622 \tiny{$\pm$ 0.0026} & 0.6753 $\pm$ 0.0028 & 0.6079 $\pm$ 0.0036 & 0.6848 $\pm$ 0.0074 & 0.7273 $\pm$ 0.0054 & 0.7793 $\pm$ 0.0055 & 0.9364 \tiny{$\pm$ 0.0008} & 0.8674 \tiny{$\pm$ 0.0036} \\

+$\gca$[0]-sigmoid,4 & 0.7257 \tiny{$\pm$ 0.0016} & 0.7934 $\pm$ 0.0015 & 0.8044 \tiny{$\pm$ 0.0012} & 0.8132 $\pm$ 0.0010 & 0.7257 $\pm$ 0.0016 & 0.8269 $\pm$ 0.0015 & 0.8611 $\pm$ 0.0013 & 0.8959 $\pm$ 0.0013 & 0.5968 \tiny{$\pm$ 0.0059} & 0.6422 $\pm$ 0.0030 & 0.6554 \tiny{$\pm$ 0.0032} & 0.6696 $\pm$ 0.0039 & 0.5968 $\pm$ 0.0059 & 0.6825 $\pm$ 0.0036 & 0.7238 $\pm$ 0.0047 & 0.7800 $\pm$ 0.0059 & 0.9369 \tiny{$\pm$ 0.0009} & 0.8680 \tiny{$\pm$ 0.0025} \\

+$\gca$-sigmoid,8 & 0.7283 \tiny{$\pm$ 0.0027} & 0.7943 $\pm$ 0.0017 & 0.8052 \tiny{$\pm$ 0.0014} & 0.8143 $\pm$ 0.0014 & 0.7283 $\pm$ 0.0027 & 0.8258 $\pm$ 0.0013 & 0.8595 $\pm$ 0.0009 & 0.8956 $\pm$ 0.0022 & 0.5977 \tiny{$\pm$ 0.0054} & 0.6419 $\pm$ 0.0033 & 0.6560 \tiny{$\pm$ 0.0046} & 0.6689 $\pm$ 0.0039 & 0.5977 $\pm$ 0.0054 & 0.6814 $\pm$ 0.0027 & 0.7253 $\pm$ 0.0059 & 0.7765 $\pm$ 0.0062 & 0.9364 \tiny{$\pm$ 0.0009} & 0.8655 \tiny{$\pm$ 0.0038} \\

+$\gca$[0]-tanh,4 & 0.7248 \tiny{$\pm$ 0.0027} & 0.7914 $\pm$ 0.0019 & 0.8024 \tiny{$\pm$ 0.0017} & 0.8117 $\pm$ 0.0019 & 0.7248 $\pm$ 0.0027 & 0.8250 $\pm$ 0.0019 & 0.8590 $\pm$ 0.0015 & 0.8958 $\pm$ 0.0012 & 0.6006 \tiny{$\pm$ 0.0020} & 0.6433 $\pm$ 0.0023 & 0.6569 \tiny{$\pm$ 0.0037} & 0.6699 $\pm$ 0.0031 & 0.6006 $\pm$ 0.0020 & 0.6815 $\pm$ 0.0040 & 0.7238 $\pm$ 0.0097 & 0.7754 $\pm$ 0.0067 & 0.9368 \tiny{$\pm$ 0.0002} & 0.8671 \tiny{$\pm$ 0.0042} \\

+$\gca$[0]-tanh,8 & 0.7233 \tiny{$\pm$ 0.0033} & 0.7913 $\pm$ 0.0025 & 0.8018 \tiny{$\pm$ 0.0019} & 0.8110 $\pm$ 0.0022 & 0.7233 $\pm$ 0.0033 & 0.8259 $\pm$ 0.0012 & 0.8586 $\pm$ 0.0016 & 0.8950 $\pm$ 0.0015 & 0.5986 \tiny{$\pm$ 0.0018} & 0.6441 $\pm$ 0.0021 & 0.6576 \tiny{$\pm$ 0.0021} & 0.6708 $\pm$ 0.0031 & 0.5986 $\pm$ 0.0018 & 0.6841 $\pm$ 0.0050 & 0.7258 $\pm$ 0.0054 & 0.7787 $\pm$ 0.0084 & 0.9365 \tiny{$\pm$ 0.0007} & 0.8668 \tiny{$\pm$ 0.0026} \\

+$\gca$[1]-sigmoid,4 & 0.7237 \tiny{$\pm$ 0.0026} & 0.7897 $\pm$ 0.0018 & 0.8007 \tiny{$\pm$ 0.0016} & 0.8104 $\pm$ 0.0013 & 0.7237 $\pm$ 0.0026 & 0.8247 $\pm$ 0.0022 & 0.8591 $\pm$ 0.0021 & 0.8972 $\pm$ 0.0031 & 0.6101 \tiny{$\pm$ 0.0036} & 0.6553 $\pm$ 0.0035 & 0.6699 \tiny{$\pm$ 0.0030} & 0.6832 $\pm$ 0.0029 & 0.6101 $\pm$ 0.0036 & 0.6959 $\pm$ 0.0049 & 0.7411 $\pm$ 0.0035 & 0.7935 $\pm$ 0.0039 & 0.9371 \tiny{$\pm$ 0.0007} & 0.8735 \tiny{$\pm$ 0.0030} \\

+$\gca$[1]-sigmoid,8 & 0.7221 \tiny{$\pm$ 0.0030} & 0.7885 $\pm$ 0.0028 & 0.8000 \tiny{$\pm$ 0.0024} & 0.8094 $\pm$ 0.0024 & 0.7221 $\pm$ 0.0030 & 0.8234 $\pm$ 0.0022 & 0.8592 $\pm$ 0.0024 & 0.8964 $\pm$ 0.0020 & 0.6072 \tiny{$\pm$ 0.0035} & 0.6553 $\pm$ 0.0040 & 0.6704 \tiny{$\pm$ 0.0029} & 0.6828 $\pm$ 0.0030 & 0.6072 $\pm$ 0.0035 & 0.6979 $\pm$ 0.0063 & 0.7446 $\pm$ 0.0021 & 0.7940 $\pm$ 0.0035 & 0.9365 \tiny{$\pm$ 0.0008} & 0.8730 \tiny{$\pm$ 0.0025} \\

+$\gca$[1]-tanh,4 & 0.7171 \tiny{$\pm$ 0.0000} & 0.7836 $\pm$ 0.0000 & 0.7955 \tiny{$\pm$ 0.0000} & 0.8053 $\pm$ 0.0000 & 0.7171 $\pm$ 0.0000 & 0.8192 $\pm$ 0.0000 & 0.8564 $\pm$ 0.0000 & 0.8952 $\pm$ 0.0000 & 0.6038 \tiny{$\pm$ 0.0000} & 0.6514 $\pm$ 0.0000 & 0.6681 \tiny{$\pm$ 0.0000} & 0.6807 $\pm$ 0.0000 & 0.6038 $\pm$ 0.0000 & 0.6950 $\pm$ 0.0000 & 0.7470 $\pm$ 0.0000 & 0.7975 $\pm$ 0.0000 & 0.9355 \tiny{$\pm$ 0.0000} & 0.8771 \tiny{$\pm$ 0.0000} \\

+$\gca$[1]-tanh,4 & 0.7247 \tiny{$\pm$ 0.0018} & 0.7895 $\pm$ 0.0010 & 0.8013 \tiny{$\pm$ 0.0006} & 0.8105 $\pm$ 0.0007 & 0.7247 $\pm$ 0.0018 & 0.8224 $\pm$ 0.0009 & 0.8589 $\pm$ 0.0013 & 0.8955 $\pm$ 0.0013 & 0.6077 \tiny{$\pm$ 0.0037} & 0.6522 $\pm$ 0.0037 & 0.6670 \tiny{$\pm$ 0.0033} & 0.6794 $\pm$ 0.0026 & 0.6077 $\pm$ 0.0037 & 0.6927 $\pm$ 0.0055 & 0.7384 $\pm$ 0.0047 & 0.7877 $\pm$ 0.0020 & 0.9360 \tiny{$\pm$ 0.0011} & 0.8717 \tiny{$\pm$ 0.0022} \\

+$\gca$[1]-tanh,8 & 0.7231 \tiny{$\pm$ 0.0026} & 0.7884 $\pm$ 0.0019 & 0.7996 \tiny{$\pm$ 0.0020} & 0.8092 $\pm$ 0.0018 & 0.7231 $\pm$ 0.0026 & 0.8230 $\pm$ 0.0011 & 0.8575 $\pm$ 0.0017 & 0.8956 $\pm$ 0.0017 & 0.6098 \tiny{$\pm$ 0.0037} & 0.6552 $\pm$ 0.0024 & 0.6691 \tiny{$\pm$ 0.0018} & 0.6828 $\pm$ 0.0026 & 0.6098 $\pm$ 0.0037 & 0.6969 $\pm$ 0.0039 & 0.7399 $\pm$ 0.0035 & 0.7939 $\pm$ 0.0054 & 0.9363 \tiny{$\pm$ 0.0007} & 0.8738 \tiny{$\pm$ 0.0023} \\

\bottomrule
\end{tabular}
}
\caption{LLM4CDSR abd LLM4CDSR$+\gca$ on Amazon Cloth-Sport dataset}
\label{tab:final_llm4cdsr_amazon_metrics}
\end{table}

\begin{table}[H]
\centering
\small
\setlength{\tabcolsep}{2pt}
\resizebox{\textwidth}{!}{%
\begin{tabular}{lcHcHHHHHcHcHHHHHcc}
\toprule

Model & NDCG@1 & NDCG@5 A & NDCG@10 A & NDCG@20 A & HR@1 & HR@5 A & HR@10 A & HR@20 A & NDCG@1 B & NDCG@5 B & NDCG@10 B & NDCG@20 B & HR@1 B & HR@5 B & HR@10 B & HR@20 B & AUC A & AUC B \\ \midrule

LLM4CDSR & 0.2101 \tiny{$\pm$ 0.0030} &  & 0.3512 \tiny{$\pm$ 0.3512} & &  &  &  & &0.1419 \tiny{$\pm$ 0.0008}  &  &  0.2608 \tiny{$\pm$ 0.0010}  &  &  & &  & & 0.7901 \tiny{$\pm$ 0.0008} & 0.7197 \tiny{$\pm$ 0.0011}  \\  

$+\gca[0,1]$-sigmoid,4& 0.2410 \tiny{$\pm$ 0.0012} & 0.3479 $\pm$ 0.0016 & 0.3800 $\pm$ 0.0011 & 0.4094 $\pm$ 0.0010 & 0.2410 $\pm$ 0.0012 & 0.4357 $\pm$ 0.0031 & 0.5352 $\pm$ 0.0018 & 0.6519 $\pm$ 0.0018 & 0.1994 $\pm$ 0.0054 & 0.2756 $\pm$ 0.0055 & 0.3035 $\pm$ 0.0049 & 0.3313 $\pm$ 0.0043 & 0.1994 $\pm$ 0.0054 & 0.3456 $\pm$ 0.0058 & 0.4325 $\pm$ 0.0049 & 0.5432 $\pm$ 0.0022 & 0.7937 $\pm$ 0.0013 & 0.7252 $\pm$ 0.0026 \\

$+\gca[0,1]$-sigmoid,8 & 0.2392 $\pm$ 0.0011 & 0.3466 $\pm$ 0.0014 & 0.3793 $\pm$ 0.0016 & 0.4082 $\pm$ 0.0015 & 0.2392 $\pm$ 0.0011 & 0.4356 $\pm$ 0.0013 & 0.5372 $\pm$ 0.0025 & 0.6519 $\pm$ 0.0019 & 0.1998 $\pm$ 0.0082 & 0.2761 $\pm$ 0.0070 & 0.3034 $\pm$ 0.0062 & 0.3312 $\pm$ 0.0057 & 0.1998 $\pm$ 0.0082 & 0.3462 $\pm$ 0.0067 & 0.4308 $\pm$ 0.0043 & 0.5417 $\pm$ 0.0041 & 0.7936 $\pm$ 0.0019 & 0.7247 $\pm$ 0.0010 \\

$+\gca[0,1]$-tanh,4 & 0.2395 $\pm$ 0.0018 & 0.3467 $\pm$ 0.0014 & 0.3792 $\pm$ 0.0014 & 0.4085 $\pm$ 0.0016 & 0.2395 $\pm$ 0.0018 & 0.4352 $\pm$ 0.0023 & 0.5358 $\pm$ 0.0016 & 0.6523 $\pm$ 0.0025 & 0.2014 $\pm$ 0.0028 & 0.2783 $\pm$ 0.0022 & 0.3053 $\pm$ 0.0015 & 0.3331 $\pm$ 0.0014 & 0.2014 $\pm$ 0.0028 & 0.3485 $\pm$ 0.0053 & 0.4328 $\pm$ 0.0044 & 0.5432 $\pm$ 0.0057 & 0.7927 $\pm$ 0.0022 & 0.7232 $\pm$ 0.0018 \\

$+\gca[0,1]$-tanh,8 & 0.2388 $\pm$ 0.0018 & 0.3465 $\pm$ 0.0010 & 0.3786 $\pm$ 0.0010 & 0.4078 $\pm$ 0.0010 & 0.2388 $\pm$ 0.0018 & 0.4358 $\pm$ 0.0016 & 0.5352 $\pm$ 0.0009 & 0.6512 $\pm$ 0.0018 & 0.2020 $\pm$ 0.0022 & 0.2792 $\pm$ 0.0020 & 0.3060 $\pm$ 0.0027 & 0.3332 $\pm$ 0.0026 & 0.2020 $\pm$ 0.0022 & 0.3501 $\pm$ 0.0026 & 0.4331 $\pm$ 0.0046 & 0.5415 $\pm$ 0.0051 & 0.7921 $\pm$ 0.0016 & 0.7235 $\pm$ 0.0015 \\

$+\gca[0]$-sigmoid,4 & 0.2375 $\pm$ 0.0021 & 0.3476 $\pm$ 0.0011 & 0.3803 $\pm$ 0.0008 & 0.4098 $\pm$ 0.0010 & 0.2375 $\pm$ 0.0021 & 0.4355 $\pm$ 0.0031 & 0.5367 $\pm$ 0.0021 & 0.6535 $\pm$ 0.0026 & 0.1862 $\pm$ 0.0056 & 0.2578 $\pm$ 0.0047 & 0.2848 $\pm$ 0.0047 & 0.3151 $\pm$ 0.0044 & 0.1862 $\pm$ 0.0056 & 0.3238 $\pm$ 0.0041 & 0.4079 $\pm$ 0.0049 & 0.5282 $\pm$ 0.0056 & 0.7947 $\pm$ 0.0021 & 0.7210 $\pm$ 0.0030 \\

$+\gca[0]$-sigmoid,8 & 0.2357 $\pm$ 0.0017 & 0.3469 $\pm$ 0.0011 & 0.3792 $\pm$ 0.0006 & 0.4085 $\pm$ 0.0011 & 0.2357 $\pm$ 0.0017 & 0.4365 $\pm$ 0.0024 & 0.5363 $\pm$ 0.0015 & 0.6525 $\pm$ 0.0023 & 0.1881 $\pm$ 0.0052 & 0.2591 $\pm$ 0.0026 & 0.2864 $\pm$ 0.0022 & 0.3162 $\pm$ 0.0015 & 0.1881 $\pm$ 0.0052 & 0.3243 $\pm$ 0.0037 & 0.4093 $\pm$ 0.0023 & 0.5277 $\pm$ 0.0058 & 0.7949 $\pm$ 0.0021 & 0.7207 $\pm$ 0.0019 \\

$+\gca[0]$-tanh,4 & 0.2378 $\pm$ 0.0011 & 0.3491 $\pm$ 0.0019 & 0.3815 $\pm$ 0.0018 & 0.4111 $\pm$ 0.0019 & 0.2378 $\pm$ 0.0011 & 0.4385 $\pm$ 0.0026 & 0.5389 $\pm$ 0.0020 & 0.6562 $\pm$ 0.0022 & 0.1861 $\pm$ 0.0035 & 0.2570 $\pm$ 0.0033 & 0.2845 $\pm$ 0.0027 & 0.3141 $\pm$ 0.0026 & 0.1861 $\pm$ 0.0035 & 0.3225 $\pm$ 0.0037 & 0.4082 $\pm$ 0.0026 & 0.5259 $\pm$ 0.0032 & 0.7970 $\pm$ 0.0018 & 0.7218 $\pm$ 0.0026 \\

$+\gca[0]$-tanh,8 & 0.2375 $\pm$ 0.0024 & 0.3488 $\pm$ 0.0021 & 0.3819 $\pm$ 0.0016 & 0.4116 $\pm$ 0.0016 & 0.2375 $\pm$ 0.0024 & 0.4371 $\pm$ 0.0021 & 0.5395 $\pm$ 0.0013 & 0.6577 $\pm$ 0.0009 & 0.1840 $\pm$ 0.0047 & 0.2563 $\pm$ 0.0031 & 0.2837 $\pm$ 0.0029 & 0.3130 $\pm$ 0.0023 & 0.1840 $\pm$ 0.0047 & 0.3233 $\pm$ 0.0043 & 0.4083 $\pm$ 0.0022 & 0.5250 $\pm$ 0.0024 & 0.7973 $\pm$ 0.0017 & 0.7213 $\pm$ 0.0010 \\

$+\gca[1]$-sigmoid,4 & 0.2440 $\pm$ 0.0015 & 0.3496 $\pm$ 0.0025 & 0.3814 $\pm$ 0.0029 & 0.4104 $\pm$ 0.0030 & 0.2440 $\pm$ 0.0015 & 0.4366 $\pm$ 0.0041 & 0.5352 $\pm$ 0.0058 & 0.6504 $\pm$ 0.0068 & 0.2056 $\pm$ 0.0031 & 0.2842 $\pm$ 0.0034 & 0.3116 $\pm$ 0.0009 & 0.3393 $\pm$ 0.0005 & 0.2056 $\pm$ 0.0031 & 0.3567 $\pm$ 0.0048 & 0.4418 $\pm$ 0.0032 & 0.5518 $\pm$ 0.0059 & 0.7916 $\pm$ 0.0046 & 0.7238 $\pm$ 0.0050 \\

$+\gca[1]$-sigmoid,8 & 0.2445 $\pm$ 0.0016 & 0.3500 $\pm$ 0.0013 & 0.3821 $\pm$ 0.0006 & 0.4118 $\pm$ 0.0007 & 0.2445 $\pm$ 0.0016 & 0.4389 $\pm$ 0.0024 & 0.5386 $\pm$ 0.0010 & 0.6563 $\pm$ 0.0020 & 0.2118 $\pm$ 0.0028 & 0.2876 $\pm$ 0.0036 & 0.3158 $\pm$ 0.0028 & 0.3429 $\pm$ 0.0028 & 0.2118 $\pm$ 0.0028 & 0.3574 $\pm$ 0.0044 & 0.4452 $\pm$ 0.0031 & 0.5529 $\pm$ 0.0020 & 0.7966 $\pm$ 0.0015 & 0.7282 $\pm$ 0.0029 \\

$+\gca[1]$-tanh,4 & 0.2427 $\pm$ 0.0024 & 0.3484 $\pm$ 0.0025 & 0.3805 $\pm$ 0.0025 & 0.4099 $\pm$ 0.0034 & 0.2427 $\pm$ 0.0024 & 0.4373 $\pm$ 0.0039 & 0.5371 $\pm$ 0.0043 & 0.6536 $\pm$ 0.0081 & 0.2097 $\pm$ 0.0033 & 0.2857 $\pm$ 0.0033 & 0.3140 $\pm$ 0.0025 & 0.3410 $\pm$ 0.0022 & 0.2097 $\pm$ 0.0033 & 0.3555 $\pm$ 0.0040 & 0.4431 $\pm$ 0.0041 & 0.5504 $\pm$ 0.0013 & 0.7951 $\pm$ 0.0048 & 0.7267 $\pm$ 0.0032 \\

$+\gca[1]$-tanh,8 & 0.2425 $\pm$ 0.0019 & 0.3488 $\pm$ 0.0028 & 0.3814 $\pm$ 0.0030 & 0.4106 $\pm$ 0.0034 & 0.2425 $\pm$ 0.0019 & 0.4365 $\pm$ 0.0040 & 0.5376 $\pm$ 0.0055 & 0.6536 $\pm$ 0.0074 & 0.2049 $\pm$ 0.0034 & 0.2835 $\pm$ 0.0029 & 0.3111 $\pm$ 0.0025 & 0.3374 $\pm$ 0.0033 & 0.2049 $\pm$ 0.0034 & 0.3557 $\pm$ 0.0033 & 0.4411 $\pm$ 0.0031 & 0.5459 $\pm$ 0.0057 & 0.7949 $\pm$ 0.0049 & 0.7254 $\pm$ 0.0047 \\

\bottomrule
\end{tabular}
}
\caption{LLM4CDSR and LLM4CDSR$+\gca$ on Amazon Electronic-Phone dataset}
\label{tab:final_llm4cdsr_elec_metrics}
\end{table}

\subsection{CDSRNP}

\begin{table}[H]
\centering
\small
\resizebox{\textwidth}{!}{%
\begin{tabular}{llllllllll}
\toprule
Model &     NDCG@5 &             NDCG@10 &             NDCG@20 &              HR@5 &               HR@10 &               HR@20 \\
\midrule

CDSRNP  & 0.0940 {\tiny $\pm$ 0.0080} & 0.1170 $\pm$ 0.0080 & 0.1426 $\pm$ 0.0110 &  0.1378 $\pm$ 0.0099 & 0.2095 $\pm$ 0.0085 & 0.3115 $\pm$ 0.0224 \\

$+\gca[0]$ & 0.0978 $\pm$ 0.0083 & 0.1209 $\pm$ 0.0094 & 0.1491 $\pm$ 0.0092 & 0.1402 $\pm$ 0.0124 & 0.2115 $\pm$ 0.0153 & 0.3244 $\pm$ 0.0184 \\

$+\gca[01]$ & 0.0983 $\pm$ 0.0070 & 0.1226 $\pm$ 0.0073 & 0.1495 $\pm$ 0.0089 & 0.1402 $\pm$ 0.0101 & 0.2156 $\pm$ 0.0136 & 0.3227 $\pm$ 0.0165 \\

$+\gca[012]$ & 0.0972 $\pm$ 0.0079 & 0.1219 $\pm$ 0.0076 & 0.1483 $\pm$ 0.0084 & 0.1406 $\pm$ 0.0117 & 0.2168 $\pm$ 0.0138 & 0.3228 $\pm$ 0.0138 \\

$+\gca[0123]$ & 0.0967 $\pm$ 0.0071 & 0.1215 $\pm$ 0.0092 & 0.1474 $\pm$ 0.0081 & 0.1428 $\pm$ 0.0118 & 0.2200 $\pm$ 0.0204 & 0.3235 $\pm$ 0.0131 \\

$+\gca_{+}[0123]$ & 0.0965 $\pm$ 0.0089 & 0.1229 $\pm$ 0.0125 & 0.1461 $\pm$ 0.0096 & 0.1408 $\pm$ 0.0135 & 0.2230 $\pm$ 0.0229 & 0.3159 $\pm$ 0.0105 \\

$+\gca_{+}[012]$ & 0.0986 $\pm$ 0.0071 & 0.1232 $\pm$ 0.0083 & 0.1486 $\pm$ 0.0084 & 0.1426 $\pm$ 0.0130 & 0.2196 $\pm$ 0.0173 & 0.3211 $\pm$ 0.0159 \\

$+\gca_{\text{SH}}[0]$ & 0.0976 $\pm$ 0.0075 & 0.1215 $\pm$ 0.0084 & 0.1491 $\pm$ 0.0095 & 0.1398 $\pm$ 0.0112 & 0.2136 $\pm$ 0.0140 & 0.3240 $\pm$ 0.0209 \\

$+\gca_{\text{SH}}[01]$ & 0.0980 $\pm$ 0.0067 & 0.1219 $\pm$ 0.0074 & 0.1489 $\pm$ 0.0086 & 0.1406 $\pm$ 0.0094 & 0.2144 $\pm$ 0.0145 & 0.3219 $\pm$ 0.0156 \\

$+\gca_{\text{SH}}[012]$ & 0.0971 $\pm$ 0.0077 & 0.1218 $\pm$ 0.0078 & 0.1479 $\pm$ 0.0080 & 0.1402 $\pm$ 0.0120 & 0.2168 $\pm$ 0.0144 & 0.3212 $\pm$ 0.0131 \\

$+\gca_{\text{SH}}[0123]$ & 0.0972 $\pm$ 0.0080 & 0.1217 $\pm$ 0.0100 & 0.1477 $\pm$ 0.0095 & 0.1433 $\pm$ 0.0119 & 0.2195 $\pm$ 0.0203 & 0.3235 $\pm$ 0.0152 \\

$+\gca_{\text{SH+}}[0123]$ & 0.0980 $\pm$ 0.0072 & 0.1225 $\pm$ 0.0098 & 0.1485 $\pm$ 0.0094 & 0.1426 $\pm$ 0.0112 & 0.2192 $\pm$ 0.0221 & 0.3231 $\pm$ 0.0181 \\

$+\gca_{\text{SH+}}[012]$ & 0.0983 $\pm$ 0.0069 & 0.1226 $\pm$ 0.0087 & 0.1485 $\pm$ 0.0081 & 0.1430 $\pm$ 0.0124 & 0.2192 $\pm$ 0.0177 & 0.3228 $\pm$ 0.0155 \\
\bottomrule
\end{tabular}
}
\caption{CDSRNP and CDSRNP+$\gca$ on Amazon Electronics-Phone Dataset}
\end{table}

\end{document}